\documentclass[preprint,12pt]{elsarticle}

\usepackage{times}
\usepackage{soul}
\usepackage{url}
\usepackage[hidelinks]{hyperref}
\usepackage[utf8]{inputenc}
\usepackage[small]{caption}
\usepackage{subcaption}
\usepackage{graphicx}
\usepackage{amsmath}
\usepackage{amsthm}
\usepackage{booktabs}
\usepackage{algorithm}
\usepackage{cellspace}
\usepackage{algorithmic}
\usepackage{geometry}

\usepackage{footmisc}
\usepackage{bm}
\usepackage{todonotes}
\usepackage{amsfonts}
\usepackage{mathtools}
\usepackage{multirow}
\usepackage{enumitem} 
\usepackage{booktabs}
\usepackage{xspace}
\usepackage{comment}
\newcommand{\act}[1]{\ensuremath{\mathfrak{#1}}}
\newcommand{\odrl}{Offline Deep RL\xspace}
\newcommand{\actlabel}[1]{\textsc{#1}}

\newcommand{\rims}{$\textsc{RIMS}$\xspace}

\journal{Information Systems}

\begin{document}

\begin{frontmatter}

\title{Learning optimal policies from event logs through reinforcement learning: a comparison of deep and
MDP-based approaches}


	\author[1]{Stefano Branchi}
	\ead{sbranchi@fbk.eu}
	\author[1,2]{Andrei Buliga}
	\ead{abuliga@fbk.eu}
	\author[3]{Chiara Di Francescomarino}
	\ead{c.difrancescomarino@unitn.it}
	\author[2]{Chiara Ghidini}
	\ead{chiara.ghidini@unibz.it}
	\author[1]{Riccardo Graziosi}
	\ead{rgraziosi@fbk.eu}
	\author[1,2,4]{Francesca Meneghello}
	\ead{fmeneghello@fbk.eu}
	\author[1]{Massimiliano Ronzani }
	\ead{mronzani@fbk.eu}
	\affiliation[1]{organization={Fondazione Bruno Kessler (FBK)},
		city={Povo},
		country={Italy}}
	\affiliation[2]{organization={Free University of Bolzano},
		city={Bolzano},
		country={Italy}}
	\affiliation[3]{organization={University of Trento},
		city={Trento},
		country={Italy}}
	\affiliation[4]{organization={University of Rome, La Sapienza},
		city={Rome},
		country={Italy}}

\begin{abstract}
 Prescriptive Process Monitoring is an emerging area within Process Mining that focuses on recommending actions to optimize business outcomes.
Most existing works focus on prescribing pre-defined interventions, that is, pre-defined (sets of) actions, on specific ongoing process executions, aimed at achieving a specific objective or Key Performance Indicator (KPI).
In contrast, only a few approaches have explored the \emph{learning} and \emph{evaluation} of optimal behavioral policies, that is, general strategies that determine the best sequence of actions that constitute a process execution in order to maximize the desired KPI.

In this paper, we address the problem of \emph{learning} optimal behavioral policies by proposing an AI-based approach that learns an optimal policy directly from historical process executions using Reinforcement Learning (RL), with the goal of recommending the best actions to optimize a KPI of interest. 
To this end, we employ two distinct RL techniques. The first is a classical, model-based approach that extends previous work by the authors, overcoming its limitations, by constructing a Markov Decision Process (MDP) that captures the process behavior.
The second is a model-free technique based on offline Deep RL, a rapidly advancing family of methods that have demonstrated strong performance across a variety of domains. 
Differently from state of the art work, we aim at building methods that minimize the usage of domain knowledge on the scenario at hand, and learn optimal policies directly from historical event data.
In this way, we investigate whether, given a relevant KPI for the process under analysis, it is possible not only to learn when to apply an intervention, but also to discover which interventions are effective directly from data.
Moreover, we aim at targeting complex scenarios, such as the ones modeling an interplay with customers or external actors, in which the process owner (and therefore the behavioral policy) may control only part of the process activities. 

Concerning the \emph{evaluation}, we adopt an approach already used in the evaluation of pre-defined interventions of exploiting Business Process Simulation (BPS), but we adapt and customize it to the task of evaluating optimal behavioral policies.
In particular, we build a data-driven BPS environment, to evaluate the discovered policies.
Our results show that both methods consistently improve the targeted KPI with similar effectiveness, with the model-based approach outperforming offline Deep RL in terms of computational efficiency.

\end{abstract}


\end{frontmatter}



\section{Introduction}
\label{sec:intro}

Prescriptive Process Monitoring (PPM) is a prominent problem in Process Mining, which consists in identifying a set of actions, or interventions, to be recommended with the goal of optimising a target measure of interest or Key Performance Indicator (KPI). 
In its simplest formulations it contains methods for raising alarms or triggering pre-defined interventions~\cite{Teinemaaetal2018,fahrenkrog2019fire,metzger2019proactive,metzger2020triggering,DBLP:journals/corr/abs-2109-02894,shoush2024prescriptive}, to
prevent or mitigate undesired outcomes by recommending the best activity to perform~\cite{de2020design} or the most suitable resource allocation~\cite{padella_24,park_prediction-based_2019,10.1007/978-3-319-39696-5_35}.


A key aspect of Prescriptive Process Monitoring is then the identification of the interventions or treatments to be recommended.
This task is often non-trivial, as~\cite{Deesetal2019} shows that expert-defined interventions are not always effective, even when the policy
determining when to apply them is highly accurate. This highlights the importance of selecting effective interventions based on data.
State of the art approaches have mainly targeted the application of individual pre-defined interventions, or treatments, focusing on learning whether and when such interventions should be applied. 
Only few recent works have targeted the problem of discovering general behavioral policies. Specifically, these approaches propose methods for recommending the best next activity to perform in order to optimize a given KPI of interest, such as the process cycle time or outcome~\cite{weinzierl2020predictive,groger2014prescriptive,de2020design}.
In doing so, these methods inherently combine the exploration of possible recommendations with their exploitation.

However, a common but often simplistic assumption in many of these works is that the prescriptive model can recommend any activity occurring in the process. In many cases, the process owner typically controls only a subset of activities, while others are influenced by exogenous factors or external actors. Think for example of a process that concerns the interplay with customers or external providers. In all these scenarios the behavioral policy has to take into account the fact that it may control only part of the process activities.  

Only a few works~\cite{us@BPM22,DBLP:journals/dke/HundoganVTRL25} explicitly consider process execution in the context of a multi-actor environment influenced by exogenous factors, including the behavior of other process participants.
In such settings, identifying the best strategy for a specific target actor is non-trivial, and Artificial Intelligence (AI) techniques based on Reinforcement Learning (RL) have shown promising potential~\cite{DBLP:journals/access/KhaidemLYALD20, DBLP:conf/aaai/Micheli021,Go216,li-etal-2016-deep}.

Two main RL paradigms are commonly used in the literature:
(i) model-free approaches, which rely solely on data and typically leverage Deep Reinforcement Learning methods~\cite{DRL}; and
(ii) model-based approaches~\cite{Sutton1998}, which assume the availability of a model for ``playing the game'', typically formalized as a Markov Decision Process (MDP). 
The first approach has the advantage of avoiding the need to build explicit models; moreover, the use of Deep Learning facilitates handling large or continuous state and action spaces, though it often results in opaque decisions from the perspective of RL reasoning.
The second approach is more transparent, as it relies on an explicit model, typically a Markov Decision Process (MDP), but it requires such a model to be available or constructed for the scenario and KPI of interest. In the context of Process Mining, this is challenging because the knowledge needed to define an MDP is usually not readily available. Extracting it manually from event logs (as done in~\cite{us@BPM22}) is prone to errors and requires extensive knowledge on the domain at hand, and existing Process Discovery techniques~\cite{DBLP:books/sp/22/Aalst22a} are not designed to produce MDP states that align with relevant KPIs.


The goal of this paper is to assess and compare the capabilities of these two RL paradigms in learning an optimal policy directly from event data.
To this end, we introduce two AI-based approaches, each instantiating one of the two paradigms, to learn an optimal policy that recommends the best activities to perform in order to optimize a KPI of interest in a multi-actor setting.
Both approaches exploit data, specifically sequences of past process executions annotated with minimal knowledge about the assignment of activities to actors and the KPI of interest.
In particular, the model-based approach uses this information to build an MDP that reliably models the process environment.

To this end, we explicitly consider multi-actor processes, distinguishing between activities controlled by the target actor and those performed by external participants. As for the interventions, the proposed approaches do not rely on extensive domain knowledge to manually exclude candidate interventions and limit treatments to pre-defined ones. Instead, all controllable activities are initially considered, as the RL process naturally discourages recommendations that do not provide a net benefit with respect to the target KPI, effectively filtering out ineffective or overly costly actions.




To demonstrate the validity of our approach, we leverage Business Process Simulation (BPS).
The idea of applying Business Process Simulation (BPS) to learn and evaluate recommendation policies was first introduced in the context of resource allocation~\cite{decision_support_dumas07,DBLP:journals/dke/HuangALD11,padella_24,Meneghello_bpm,DBLP:journals/dke/HuangALD11}, as the resource perspective is one of the easiest aspects to modify within a simulation model.
More recently, BPS has also been used to evaluate prescriptions related to other process perspectives~\cite{Weytjens_CI_vs_RL}, but these approaches are limited to intervention-based recommendations and are not directly applicable to the evaluation of more general policies.
Therefore we specifically design a BPS model capable of incorporating and evaluating next-activity selection policies. Our evaluation is split in two parts: first, we use a synthetic process model to generate event logs and evaluate the learned policies. This allows us to explore how different statistical properties of the generated logs affect the resulting policies; second, we start from real-world event logs and build the hybrid BPS model from data \cite{simod,rims}, which allows us to accurately replicate the process behavior. This enables us to test the applicability and effectiveness of our approach in realistic domains.

The results show that both proposed methods reliably learn effective policies that consistently improve the targeted KPI across real and synthetic scenarios, with respect to the baseline policy followed in the historical process executions. The comparison between the model-based and model-free techniques indicates a slight advantage for the former in terms of performance and a substantial gain in terms of computational efficiency.

The paper is structured as follows:
Section~\ref{sec:motivating_example} provides a motivating example of a synthetic scenario to clarify the challenge addressed by the proposed approaches;
Section~\ref{sec:related} reviews related works;
Section~\ref{sec:background} introduces the core technical background;
Section~\ref{sec:learning} presents the two RL methods proposed for learning policies directly from event data;
Section~\ref{sec:evaluation} defines the evaluation settings; 
Section~\ref{sec:result} presents and discusses the evaluation results by also exploring its limitations and, finally,
Section~\ref{sec:conclusion} concludes the paper offering possible future directions.


\section{Motivating Example}
\label{sec:motivating_example}

In this section, we present a running example based on a synthetic scenario to clarify the type of problem we aim to solve and the characteristics of the multi-actor processes we address.

\begin{figure*}[t]
	\centering
	\includegraphics[width=\textwidth]{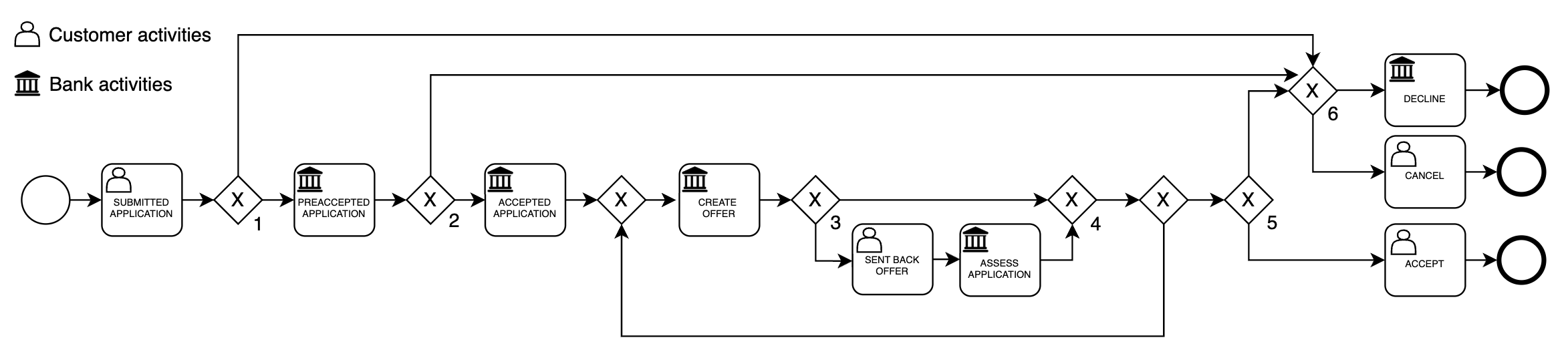} 
	\caption{BPMN model describing the loan application process.}
    \label{fig:example}
\end{figure*}

\begin{table}[ht]
\centering
\resizebox{1\columnwidth}{!}{
\begin{tabular}{clccl}
\toprule
Case ID & Activity & Timestamp & Requested Amount & Ownership \\
\midrule
1001 & \actlabel{submitted application}     & 2025-05-02 09:15 & €12,000 & Customer \\
1001 & \actlabel{preaccepted application}   & 2025-05-02 14:30 & €12,000 & Bank     \\
1001 & \actlabel{accepted application}   & 2025-05-02 15:30 & €12,000 & Bank     \\
1001 & \actlabel{create offer}              & 2025-05-03 10:00 & €12,000 & Bank     \\
1001 & \actlabel{sent back offer}           & 2025-05-06 17:45 & €12,000 & Customer \\
1001 & \actlabel{assess application}        & 2025-05-07 09:00 & €12,000 & Bank     \\
1001 & \actlabel{accept}                    & 2025-05-07 09:10 & €12,000 & Customer \\
\midrule
1002 & \actlabel{submitted application}      & 2025-05-03 08:50 & €20,000 & Customer \\
1002 & \actlabel{preaccepted application}   & 2025-05-03 16:20 & €20,000 & Bank     \\
1002 & \actlabel{accepted application}    & 2025-05-04 10:00 & €20,000 & Bank     \\
1002 & \actlabel{create offer}               & 2025-05-04 12:10 & €20,000 & Bank     \\
1002 & \actlabel{cancel}                    & 2025-05-08 15:10 & €20,000 & Customer \\
\bottomrule
\end{tabular}
}
\caption{Excerpt from the event log of the synthetic loan application process, including annotations of activity ownership.}
\label{tab:example_log}
\end{table}

Consider a loan application process similar to those featured in the BPI Challenges 2012~\cite{vandongen_2012} and 2017~\cite{bpic17}.
Figure~\ref{fig:example} shows a BPMN model\footnote{https://www.bpmn.org/} for a simplified version of such a process.
Each process execution corresponds to a single loan application submitted by a customer to the bank (\actlabel{submitted application}).
The application can be accepted (\actlabel{preaccepted application}, \actlabel{accepted application}) or rejected (\actlabel{decline}) by the bank. Similarly, the customer can withdraw the application at any time (\actlabel{cancel}).
If the bank decides to grant the loan, it generates and sends one or more offers to the customer (\actlabel{create offer}).
The customer may respond to the bank (\actlabel{sent back offer}), after which the bank assesses the customer’s intention (\actlabel{assess application}) to either accept the offer (\actlabel{accept}) or refuse it (\actlabel{cancel}).
An example of traces generated from this process model is reported in Table~\ref{tab:example_log}.

This process is an example of a multi-actor scenario involving two main actors: the \emph{bank} and the \emph{customer}.
Each actor controls certain activities and pursues their own objective: the customer aims to obtain a convenient loan offer, whereas the bank seeks reliable customers to whom it can issue loan offers.

Let us consider the perspective of one of the two actors. In this work, we focus on the bank’s perspective, as it is the most relevant from a Business Process Management (BPM) point of view and corresponds to the entity that collects and owns the execution data in real-world settings.

The bank’s overall goal is for the customer to accept the offer, an outcome represented by the activity \actlabel{accept}.
If the customer declines the offer, the time and resources the bank invested in generating it are effectively wasted.

We aim to prescribe what the bank should do, for each process instance, in order to maximize this objective.
This involves suggesting which activity is best to perform at each point in time. A key caveat, however, is that not all activities are under the bank’s control: some are triggered by customer actions (as indicated by the icons in Figure~\ref{fig:example}) and therefore cannot be directly prescribed. Notably, the \actlabel{accept} activity, which represents the bank’s main objective, is ultimately decided by the customer.
The main assumption of this work
is that the policy adopted by the bank in selecting its actions can influence the customer’s behavior and response, potentially increasing the likelihood that the customer accepts the loan offer.

This type of problem naturally fits within the Reinforcement Learning (RL) framework, where the bank represents the agent for which we aim to learn a policy, and the customers correspond to the environment that responds to the bank’s actions.

The main challenge in applying RL techniques to this setting is the impracticality of interacting directly with the real business process to learn and assess the policy.
Exploration in a live system can be costly, disruptive, and potentially risky.
Before deploying a prescriptive policy in practice, process stakeholders typically require some guarantees regarding its effectiveness and safety.

To address this challenge, in this paper we propose a method to learn a policy using RL directly from historical event data (Table~\ref{tab:example_log}), and outline how data-driven business process simulation can support its assessment prior to deployment.

\section{Related work}
\label{sec:related}

This section reviews related work in the field of Prescriptive Process Monitoring, with a particular focus on best-next-activity recommendation and on the evaluation of prescriptive methods.

Prescriptive Process Monitoring refers to a set of techniques in Process Mining aimed at recommending actions during process execution to improve outcomes such as Key Performance Indicators (KPIs).
A review of techniques from the early phase of Prescriptive Process Monitoring can be found in~\cite{DBLP:journals/corr/abs-2112-01769}.
Focusing on the type of recommended interventions, existing approaches can be broadly categorized into two main groups.
The first group includes methods that recommend predefined interventions aimed at preventing or mitigating undesired outcomes~\cite{Teinemaaetal2018,fahrenkrog2019fire,metzger2019proactive,metzger2020triggering,DBLP:journals/corr/abs-2109-02894,shoush2024prescriptive}.
In this work, we focus on the second group of approaches, which aim to learn general behavioral policies that recommend the next activity to perform in order to achieve a desired outcome. In a sense, this approach inherently includes the discovery of the intervention itself.

The work in~\cite{weinzierl2020predictive, DBLP:conf/bpm/WeinzierlDZM20} discusses how the most likely behavior does not guarantee to achieve the desired business goal. As a solution to this problem, the authors propose and evaluate a prescriptive business process monitoring technique that recommends next best actions to optimize a specific KPI, i.e., the time.
%
The work in~\cite{groger2014prescriptive} uses prescriptive analytics to proactively generate action recommendations during execution in a manufacturing scenario, with the goal of preventing predicted deviations in key performance metrics.
The action prescribed is not limited to the next activity to perform but include values or conditions on its data payload.
The work in~\cite{de2020design} discusses Process-Aware Recommender (PAR) systems, where a prescriptive analytics component, leveraging a transition system, suggests next activities for ongoing executions predicted to have a negative outcome, with the aim of reducing the risk of undesirable process completion.
A common limitation of both these approaches is that they allow the recommendation of any subsequent activity during process execution, whereas in many real-world settings the process owner has control over only a subset of the activities.

Only a few approaches have formulated the RL setting for complex processes where the agent can perform only a subset of activities. For example, in~\cite{us@BPM22}, the authors model the agent’s limited control---restricting decisions to its scope---while accounting for the influence of other actors on the overall process. They build an explicit MDP to capture the process behavior and use it to train an RL agent.
A key limitation of this approach is its reliance on domain knowledge to define the state space effectively, making it less adaptable to new domains.

The most recent work in this direction~\cite{DBLP:journals/dke/HundoganVTRL25} uses RL techniques for next-activity recommendations in care processes. A multi-actor process setting is also required here. The RL problem is defined with states representing patient-initiated incidents (e.g., verbal or physical aggression, self-harm) and actions as staff countermeasures (e.g., talking, distracting, or secluding the patient), with stochastic patient responses. The authors construct an MDP to model the process and train an RL agent, evaluating the learned policy against simple heuristics through simulations based on the same MDP.

Both~\cite{us@BPM22} and~\cite{DBLP:journals/dke/HundoganVTRL25} share a common approach mismatch: they construct an explicit MDP to model the process, yet rely on model-free RL techniques~\cite{Sutton1998}, such as Monte Carlo methods in~\cite{us@BPM22}, and SARSA and Q-learning in~\cite{DBLP:journals/dke/HundoganVTRL25}, which are designed to operate when an explicit model is not available.
While this combination can be justified when using the MDP as a simulator to support learning in complex environments, it introduces a conceptual redundancy and may lead to unnecessary complexity or inefficiencies.
When an explicit MDP is available, model-based methods such as dynamic programming are typically preferable, as they fully exploit the model and are computationally more efficient.

In this paper, we aim to address this gap by adopting the perspective of a single process actor in a complex multi-actor process scenario, following~\cite{us@BPM22}, and optimizing a domain-specific KPI of interest through a Reinforcement Learning (RL) approach. However, the method proposed in~\cite{us@BPM22} relies heavily on background knowledge and manual preprocessing to define the state space of the RL problem.
In contrast, this work aims to build the RL pipeline automatically, starting solely from the event log, the KPI of interest, and the activity ownership, i.e., the distinction between activities controlled by the process owner, which can be recommended, and those influenced by external factors, which must be properly accounted for.

Additionally, all the approaches reviewed in this section evaluate the learned policies by performing an analysis on a test event log.
However, as we will discuss later, this type of evaluation is inherently biased, since the event log reflects the behavior induced by the original policy under which the data was generated, which is usually different from the one being tested.
In this paper, we address this limitation by employing Business Process Simulation, which enables a form of controlled A/B testing on the simulated process. This allows for a fairer comparison between the performance of the learned policy and that of the original policy used in reality.



\section{Background}
\label{sec:background}

In this section we provide the background knowledge necessary to understand the rest of the paper.

\subsection{Event log}
\label{sec:back_event_log}

An \emph{event log} $\mathcal{L}$ consists of traces representing executions of a process (also known as \emph{cases}).
A \emph{trace} $\sigma = \langle e_1, e_2, \dots, e_n \rangle$ is a finite, temporally ordered sequence of events $e_i$,
each referring to the execution of an activity label $\act{a} \in A$.
The \emph{event} $e_i = (\act{a},t, p_1,\ldots,p_m)$ is, in turn, a complex structure comprising the activity label $\act{a}=\text{Act}(e_i)$, its timestamp $t$, indicating the time in which the event has occurred, and possibly \emph{data payloads} $p_1,\ldots,p_m$, consisting of categorical and numeric attributes, such as, the resource(s) involved in the execution of an activity, or other data recorded during the event.
\emph{Data payloads} can also be associated with the trace itself and remain constant throughout its execution, unlike those related to individual events that compose the trace.
A partial trace execution, or \emph{prefix}, of a trace $\sigma = \langle e_1, e_2, \dots,
e_k \rangle$ is a sequence $\sigma_k$ containing the first $k$ events of $\sigma$, with $k \le n$.\footnote{We allow $k = n$, meaning that a complete trace can be considered a prefix of itself. While this slightly extends the conventional notion of a prefix, it will simplify the definition of MDP states in Section~\ref{sec:learning}.}
For instance, given a trace $\sigma = \langle e_1, e_2, e_3, e_4 \rangle$, the prefix $\sigma_2 = \langle e_1, e_2 \rangle$.

We define the \emph{$\act{a}$-frequency} $f_{\act{a}}(\sigma)$ of a trace $\sigma$, 
the number of times the activity label $\act{a}$ occurs in the trace $\sigma$.
We define the \emph{$\act{a}$-position} $p_{\act{a}}(\sigma)$ of a trace $\sigma$, 
the \emph{last position} in which the activity label $\act{a}$ occurs in the trace $\sigma$,
with the first event having position 1.
Both frequency and position are equal to zero if the corresponding activity is missing in the trace.
For instance, given a trace $\sigma = \langle (\act{a}_1, t_1), (\act{a}_2, t_2), (\act{a}_1, t_3)\rangle$, $f_{\act{a}_1}(\sigma)=2$ and $p_{\act{a}_1}(\sigma)=3$.

\subsection{Reinforcement Learning}
\label{sec:background_RL}

Reinforcement Learning~\cite{Sutton1998}
is an interactive learning paradigm where an agent learns  to act in an environment by
maximizing the total amount of reward received by its actions.
At each time step $t$, the agent chooses and performs an \emph{action} $a$ in response to the observation of the \emph{state} of the environment $s$.
Performing action $a$ causes, at the next time step $t+1$, the environment to stochastically move to a new state $s'$,
and gives the agent a \emph{reward} $r_{t+1}=\mathcal{R}(s,a,s')$ that indicates how well the agent has performed.
The probability that, given the current state $s$ and the action $a$, the environment moves into the new state $s'$
is given by the state transition function $P(s,a,s')$.

The learning problem is therefore described as a discrete-time MDP $M$, 
which is defined by a tuple $M=(\mathcal{S}, \mathcal{A}, P,\mathcal{R}, \gamma)$, where:
\begin{itemize}
    \item $\mathcal{S}$ is the set of states.
    \item $\mathcal{A}$ is the set of agent's actions.
    \item $P:\mathcal{S} \times \mathcal{A} \times \mathcal{S} \rightarrow \mathbb{R}_{[0,1]}$ is the transition probability function.
		$P(s,a,s') = Pr(s_{t+1} = s' \mid s_t = s, a_t = a)$ is the probability of transition (at the time step $t$) from state $s$ to state $s'$
		under action $a \in \mathcal{A}$.
    \item $\mathcal{R}: \mathcal{S} \times \mathcal{A} \times \mathcal{S} \rightarrow \mathbb{R}$ is the reward function.
		$\mathcal{R}(s,a,s')$ is the reward obtained by going from $s$ to $s'$ with action $a$.
    \item $\gamma \in \mathbb{R}_{[0,1]}$ is the discount factor for the rewards.
		Value of $\gamma < 1$  models an agent that discounts future rewards.
\end{itemize}
An MDP satisfies the \emph{Markov Property}, that is, given $s_t$ and $a_t$, the next state $s_{t+1}$ is conditionally independent
of all prior states and actions, and it only depends on the current state, i.e., $Pr(s_{t+1}\mid s_t, a_t)=Pr(s_{t+1}\mid s_0, \cdots, s_t, a_0, \cdots, a_t)$.
A \emph{deterministic policy}\footnote{In this work, we focus on deterministic policies. However all definitions presented in this section can be generalized to stochastic policies.} $\pi: \mathcal{S} \rightarrow \mathcal{A}$ is a mapping from each state $s \in \mathcal{S}$ to an action $a \in \mathcal{A}$,
and the \emph{cumulative reward} is the (discounted) sum of the rewards obtained by the agent while acting at the various time points. 
The \emph{state-action value function} $q^\pi$ is the expected discounted future reward obtained by taking action $a$ in state $s$ and then continuing to use the policy $\pi$ thereafter
\begin{equation} 
\label{eq:q_value}
	q^{\pi}(s,a) := \mathbb{E}_\pi
	\bigg(\sum_{k=0}^\infty\gamma^k r_{k+t+1} \bigg|s=s_t,a=a_t \bigg)
\end{equation}
where $r_{t+1} = \mathcal{R}(s_t,a_t,s_{t+1})$ is the reward obtained at time $t+1$ and $\mathbb{E}$ means expectation value.
The state-action value function satisfies the following recursive relation called \emph{Bellman equation}
\begin{equation}
\label{eq:Bellman}
	q^{\pi}(s,a) = \sum_{s'} P(s,a,s') \left( \mathcal{R}(s,a,s') + \gamma\, q^{\pi}(s',a') \right)
\end{equation}
where $a'=\pi(s')$ is the action selected by the deterministic policy $\pi$ at the state $s'$.
A policy is \emph{optimal}, usually denoted as $\pi^*$, if it maximizes the state-action value~\eqref{eq:q_value},
that is if $q^*(s,a) \coloneqq q^{\pi^*}(s,a)\ge q^{\pi}(s,a)$ for all $\pi: \mathcal{S}\to\mathcal{A}$, $s\in \mathcal{S}$ and $a\in \mathcal{A}$.
Learning the optimal policy is the goal of RL, and can be achieved exploiting the \emph{Bellman optimality equation}
\begin{equation}
\label{eq:Bellman_opt}
	q^*(s,a) = \sum_{s'} P(s,a,s') \left( \mathcal{R}(s,a,s') + \gamma \max_{a'} q^*(s', a') \right).
\end{equation}

In this paper, we use a Dynamic Programming (DP) algorithm called \emph{value iteration} to solve this RL problem.
This approach is based on the recursive Bellman optimality equation~\eqref{eq:Bellman_opt}, which is applied as an update rule to iteratively refine the action-value function $q$. 
This produces a sequence of $q^{(k)}$ that converges to $q^*$ when $k\to\infty$. 
The optimal policy is then obtained by taking the greedy policy with respect to $q$ that is $\pi^*(s)=\text{argmax}_a\ q^*(s,a)$. Since value iteration is a model-based technique, it requires access to a complete model of the MDP, including the state transition probabilities that describe the environment’s dynamics. This model enables planning over the state space and searching for the optimal policy by interacting with the environment (i.e., the MDP), and thus can be considered an online RL approach.

\paragraph{\odrl}

As an alternative to the MDP-based RL approach in this paper we also employ an \odrl technique \cite{ORL_review_2020}.
This is a model-free method that does not require the construction of the MDP model, nor does it require any interaction with the environment. Instead, they are trained in a completely offline way by using historical data 
collected using a different policy, called the \emph{behaviour policy.}
This is possible through two main specific characteristics: the use of a \emph{deep neural network} as an approximation function for the state-action values $q$
and the implementation of technical precautions needed to stabilize the learning and obtain reliable policies.
\odrl has been evolving very rapidly recently since it allows to transform increasingly common large datasets of previously collected experiences
into decision making policies, in contexts where online interaction is impractical, dangerous, or extremely costly (e.g.~healthcare, production systems, autonomous driving).
In this work, as the \odrl technique, we leverage Conservative Q-Learning (CQL) \cite{CQL}, considered a state-of-the-art \odrl technique \cite{ORL_survey_2023}. The main feature of CQL is its use of a conservative q-value approximation that gives a lower bound of the policy estimated value.

\subsection{Business Process Simulation}
\label{sec:background_bps}

A BPS model denoted as $\mathcal{M} = (\mathcal{N} , \mathcal{P)}$, is
composed of two components: a business process model $\mathcal{N}$  (e.g., a BPMN model), and a set of simulation parameters $\mathcal{P}$ needed to run the simulation across all process perspectives. Specifically, the simulation parameters correspond to the resource $\mathcal{P}_R$, the time $\mathcal{P}_T$, the control flow $\mathcal{P}_C$, the inter-arrival time $\mathcal{P}_S$ and the data payload perspective $\mathcal{P}_D$, respectively.

$\mathcal{P}_R$ manages the resources involved, which are grouped into roles, and specifies, for each activity in~$\mathcal{N}$, the role(s) that can perform it.
$\mathcal{P}_T$ specifies the processing time required to complete each activity in $\mathcal{N}$. When decision points are present in $\mathcal{N}$, $\mathcal{P}_C$ governs the selection among alternative paths; otherwise, trace execution follows the standard semantics of $\mathcal{N}$.
Finally, $\mathcal{P}_{S}$ defines case inter-arrival time, which determines when the traces start the simulation, and $\mathcal{P}_{D}$ defines the 
data payloads for events and/or traces. 

BPS models are discrete event simulation (DES) models in which multiple traces are generated concurrently based on the specified inter-arrival times, and compete for resources to progress through the process. An activity within each trace can be executed only if at least one eligible resource, as defined by $\mathcal{P}_R$, is available; otherwise, the activity waits until the required resource becomes free. The resource model $\mathcal{P}_R$ also includes calendars that define the weekly availability of each resource. Accordingly, a resource can perform an activity only if it both satisfies the association rules in $\mathcal{P}_R$ and is available on its calendar. Waiting times observed during simulation result from resource contention and limited availability. The output of the simulation model $\mathcal{M}$, is a simulated event log composed of the generated traces.

To demonstrate the effectiveness of the RL approaches, we leverage BPS to perform A/B testing on the recommendations generated by the reinforcement learning (RL) techniques discussed above. Rather than relying solely on traditional test-set-based evaluation, this approach allows us to assess the impact of recommendations in the context of the full process execution. That is, recommendations are evaluated not just on isolated traces, but within the broader dynamics of the event log as a whole.


\section{Learning best next activity policy with RL}
\label{sec:learning}

In this section, we describe the two proposed methods for learning a policy that recommends the optimal next activity or set of activities directly from event data.
In Section~\ref{sec:learning_MDP}, we present the model-based RL approach, which involves constructing an MDP and training an agent using Dynamic Programming techniques.
In Section~\ref{sec:learning_odrl}, we describe the adaptation of a model-free \odrl method, namely Conservative Q-Learning (CQL), to the process mining setting.

\subsection{MDP-based RL method}
\label{sec:learning_MDP}

\begin{figure}[t]
	\centering
	\includegraphics[width=1\textwidth]{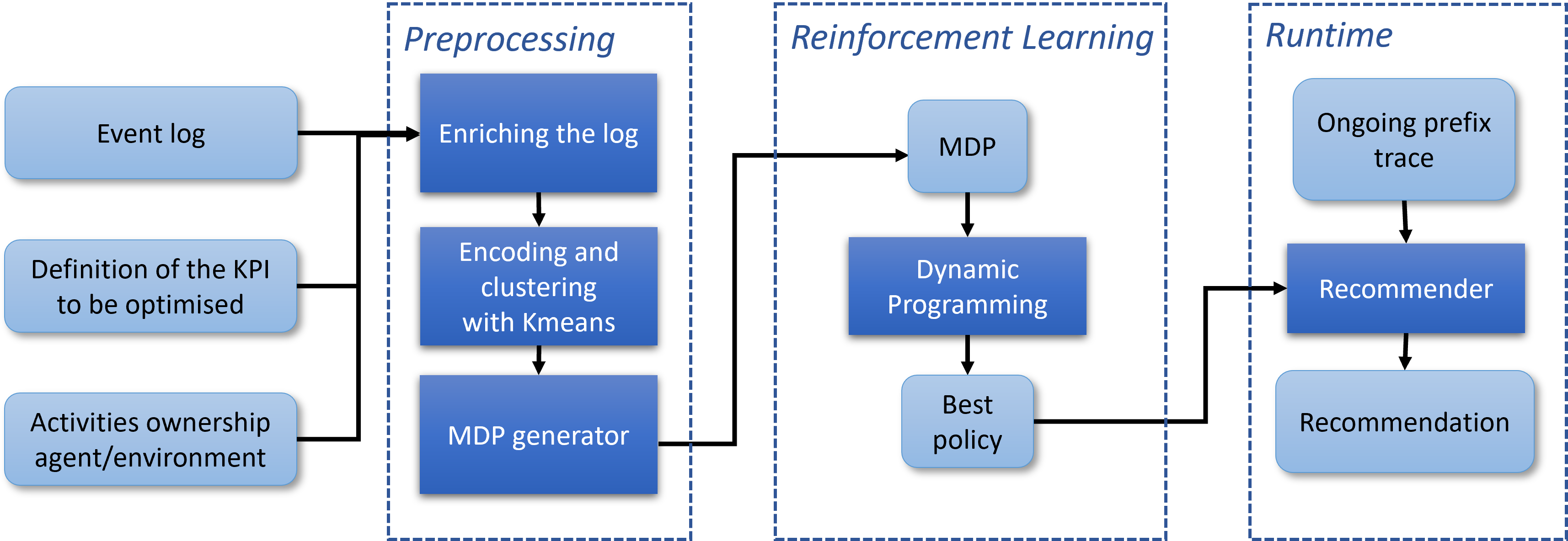} 
	\caption{Pipeline of the MDP based RL method.} \label{fig:MDP_pipeline}
\end{figure}

The pipeline for this method is shown in \figurename~\ref{fig:MDP_pipeline}.
The main idea is to mine an MDP model from an event log, so to apply Dynamic Programming (DP) methods and train an RL agent.
The pipeline takes as input an event log related to a multi-actor process, the definition of a relevant KPI to optimize, as well as the ownership of the activities in the log, and builds via RL a system that recommends the optimal next activity or set of activities in order to optimize the desired KPI.

It is composed of three phases: in the \emph{Preprocessing phase} an MDP is built starting from the three inputs of the pipeline (Section.~\ref{sec:constructionMDP}); in the \emph{Reinforcement Learning phase}
the policy is learned from the MDP by the DP value iteration algorithm
and, finally, in the \emph{Runtime phase}, the policy is used to recommend the optimal next activity or set of activities for an ongoing execution (Section~\ref{sec:RLtraining}).

\subsubsection{MDP construction from log data}
\label{sec:constructionMDP}
The aim of the \emph{Preprocessing phase} is building an MDP starting from: (i) the event log; (ii) the list of activities carried out by either the \emph{agent}, or by the \emph{environment}; and (iii) the KPI to optimize.

At a high level, the following mapping between the MDP components (actions and states) and the information extracted from the event log can be defined:
\begin{itemize}
	\item actions: the activities that the \emph{agent} can carry out; 
	\item states: a comprehensive description of the system. In a fashion similar to \cite{us@BPM22} it contains three components:
	\begin{itemize}
		\item last activity selected by the agent or the environment;
		\item the trace history, that is, the condensed information about the entire trace execution up to that point;
		\item information about the reward obtained up to that point.
	\end{itemize}
\end{itemize}

Considering the definition of MDP actions and states reported above, the steps to follow for the construction of the MDP are the following:
\begin{enumerate}
\item \emph{Enriching the log.} 
In this step, domain knowledge about activity ownership is used to label activities as either agent-controlled or environment-controlled. Agent-controlled activities are those directly under the control of the process owner, whereas environment-controlled activities are executed by external actors.
Moreover, the KPI of interest is computed for each trace $\sigma$. Its value can depend on the executed activities or on other attributes of the trace, and represents the reward $r(\sigma)$ of the whole trace $\sigma$. 
	Note that we define $r(\sigma_k)=0$ for incomplete prefixes $\sigma_{k}$, $k < \text{len}(\sigma)$. 
\item \emph{Encoding and clustering using k-means.}
	In this step, each prefix in the log
	is encoded using three types of information: the frequency of the activities, the last position in which an activity has occurred, 
	and the prefix reward.\footnote{Since the reward is non-zero only for completed traces, it effectively contributes to the state definition only for terminal states.}
	Namely, if the alphabet of all the activities is $A=\{\act{a}_1,\dots,\act{a}_n\}$
	then the encoding of a prefix $\sigma_k$ is: 
	\begin{equation}
		\bm{v}_{\sigma_k} =
		\left( \frac{f_{\act{a}_1}(\sigma_k)}{f_\text{max}}, \dots ,\frac{f_{\act{a}_n}(\sigma_k)}{f_\text{max}},
				\frac{p_{\act{a}_1}(\sigma_k)}{p_\text{max}}, \dots, \frac{p_{\act{a}_n}(\sigma_k)}{p_\text{max}}, \tilde r(\sigma_k) \right)
	\end{equation}
	where $f_{\act{a}_i}$ and $p_{\act{a}_i}$ are defined in Section~\ref{sec:back_event_log}.
	$f_\text{max}$ and $p_\text{max}$ are respectively the highest 
	frequency and position
	for all activities and all prefixes in the log. Note that the latter can also be seen as the max length of all  traces in the log
	\begin{equation}
		f_\text{max} = \max_{\substack{\act{a}\in A, \sigma \in \mathcal{L}\\ 0<k\le\text{len}(\sigma)}} \big(f_{\act{a}}(\sigma_k)\big),
		\qquad
		p_\text{max} = \max_{\sigma \in \mathcal{L}} \big(\text{len}(\sigma)\big)
	\end{equation}
	Finally, $\tilde r(\sigma_k)$ is the \emph{prefix normalized reward}: it is equal to $0$ for incomplete trace prefixes ($k<|\sigma|$), 
	while for complete traces it is the reward of the trace normalized with respect to the largest reward of all the traces. In other terms:
	\begin{equation}
	\tilde r(\sigma_k) =
	\begin{cases}
	0 & k < \text{len}(\sigma) \\
	\frac{r(\sigma) - \min_{\sigma' \in \mathcal{L}} r(\sigma')}{\max_{\sigma' \in \mathcal{L}} r(\sigma') - \min_{\sigma' \in \mathcal{L}} r(\sigma')} & k = \text{len}(\sigma)
	\end{cases}
	\end{equation}
The encoded prefixes are used to train a k-means model~\cite{kmeans}, which assigns each new prefix to the cluster containing the most similar previously observed prefixes.
Clustering serves a twofold purpose: it prevents an explosion of the dimensionality of the state space, which would otherwise increase the computational complexity of the subsequent dynamic programming procedures, and enables the MDP policy to be applied to prefixes that do not appear in the log.
The number of clusters is selected using the elbow method\footnote{We employ the KneeLocator method from the \texttt{kneed} library (\url{https://pypi.org/project/kneed/}
) with sensitivity set to 1.} applied to the Within-Cluster Sum of Squared Errors (WSS), a widely used heuristic in unsupervised learning for choosing the number of clusters, especially when no ground-truth state abstractions are available.\footnote{A more in-depth analysis of the impact of the cluster number choice for this approach is provided in~\ref{appx:cluster_number}.}
\item \emph{Constructing the MDP.}
	Once the k-means model has been fit to data, the MDP can be built.
	For each prefix $\sigma_k$, a state $s_{\sigma_k}$ is defined 
	as the pair of the last performed activity $\act{a}_{i_k}=\text{Act}(e_k)$ and the cluster $c_{\sigma_{k-1}}$ assigned to $\sigma_{k-1}$ by the k-means model:
	\begin{equation}
	\label{eq:state_def}
		s_{\sigma_k} = \big(\act{a}_{i_k},\, c_{\sigma_{k-1}}\big).
	\end{equation}
	This definition of state would allow us to include, in the state,
	information	about the history of the execution and its reward.

	Once the states are defined, the MDP is built 
	by replaying the traces in the event log: we build a directed graph, where states correspond to nodes
	and edges correspond to actions moving from one node state to another. Moreover, for each edge, the probability of reaching the target node (computed based on the
	number of traces in the event log that reach the corresponding state) and the value
	of the reward function are computed.
	Each edge is  mapped to the tuple $(s, a, s',\mathcal{P}, \mathcal{R})$ 
	where $s$ is the state corresponding to the source node of the
	edge, $a$ is the action used for labeling the edge, $s'$ is the state corresponding to the
	target node of the edge, 
    $\mathcal{P}(s, a, s')$ is computed as the percentage of traces that reach state $s'$ among all traces that reach state $s$ and execute action $a$ at the previous time step.
	The reward function $\mathcal{R}(s, a, s')$ is computed as the average of the rewards $r(\sigma_k)$
	of those prefixes $\sigma_k$ corresponding to the edge $(s, a, s')$
	\footnote{A prefix $\sigma_k$ corresponds to an edge $(s, a, s')$ if $s'=s_{\sigma_k}$, $a=\text{Act}(e_l)$ and $s=s_{\sigma_{l-1}}$,
	where $l<k$ such that $e_l$ is the last event associated to an agent's activity in $\sigma_k$.}.

\end{enumerate}

\subsubsection{RL and runtime phases}
\label{sec:RLtraining}

In the \emph{Reinforcement Learning} phase, the DP algorithm uses the MDP obtained in the previous phase to learn the optimal policy, as shown in the Reinforcement Learning box of \figurename~\ref{fig:MDP_pipeline}.
We leverage the MDP model and use value iteration method (Section.~\ref{sec:background_RL}) to iteratively compute the optimal policy through the Bellman equation \eqref{eq:Bellman_opt}.\footnote{The parameters used for the value iteration method are: discount factor $\gamma = 1$ and threshold $10^{-5}$, meaning that the algorithm stops when the maximum difference between the value functions of two successive iterations falls below the threshold.
}

In~\cite{us@BPM22}, the authors identified potential issues with MDPs mined from event data logs. Specifically, certain action choices within the MDP may result in a high final reward during training, but these choices are unreliable. This is because they occur in too few traces, meaning that the high reward is likely an artifact of the limited data, rather than a result of a robust, statistically significant correlation. As a result, this behavior is comparable to overfitting, where the MDP favors action choices that are too specific to the training data and may not generalize well to new situations.

To tackle this issue, in \cite{us@BPM22} a recalibration of the reward was performed for every transition
with a multiplicative factor depending on the number of occurrences of the given transition in the log.
In this way, very convenient but unreliable actions were discouraged during the training with respect to convenient actions that are more frequent in the log.
In this work, 
instead of scaling the reward of each transition, we scale the $q$-value function \eqref{eq:q_value} 
as follows
\begin{equation}
\label{eq:scaled_q_value}
\tilde q(s,a) =  q(s,a) \, h(s,a)
\end{equation}
where $h: \mathcal{S}\times\mathcal{A} \to \mathbb{R}_{[0,1]}$ 
is a monotonically non-decreasing function of the number $n(s,a)$ of occurrences of the state-action pair 
in the log. 
This is a generalization of the approach used in \cite{us@BPM22} since it also applies to problems where the reward is obtained only at the end of the trace, as is the case in this paper.
The trivial choice $h^0(s,a)=1$ defines the standard $q$-value.
In this work we consider two non-trivial types for $h$ and we test their effect on the learned policy:
\begin{itemize}
\item a step function $h^\text{step}_{n_T}$ which is equal to zero for $n(s,a)$ smaller or equal than a certain threshold $n_T>0$ and is equal to one for higher values of $n (s,a)$, that is:
	\begin{equation}
    \label{eq:hstep}
	h^\text{step}_{n_T}(s,a) = 
	\begin{cases}
	0 & n(s,a) \leq n_T\\
	1 & n(s,a) > n_T
	\end{cases}
	\end{equation}
\item a sigmoid function that is a smooth monotonic function from zero to one
	\begin{equation}
    \label{eq:hsig}
	h^\text{sig}_\lambda(s,a)=1-2\frac{e^{-n(s,a)/\lambda}}{1+e^{-n(s,a)/\lambda}}
	\end{equation}
	parametrized by a real value $\lambda>0$.
\end{itemize}
For every choice of $h$, we obtain a different policy $\pi^\text{MDP}_h$ from the RL training.  
The main difference between the two functions lies in their continuity and impact on transition probabilities.  
The step function is discontinuous and affects only those transitions whose occurrence count is below the threshold parameter $n_T$.  
In contrast, the sigmoid is a continuous function that, in principle, affects all transitions in the MDP, since it always holds that $h^\text{sig} < 1$.  
However, due to its exponential nature, transitions with high occurrence counts are only significantly affected when the value of $\lambda$ is large.

Finally, in the \emph{Runtime} phase, as shown in
\figurename~\ref{fig:MDP_pipeline}, the policy is used to recommend the best next activity or set of activities for an ongoing process execution\footnote{
The recommender system takes the ongoing prefix, maps it to a state following the same steps as in Section~\ref{sec:constructionMDP}
and uses the policy to recommend the best next activity to maximize the reward (KPI).}.

\subsection{\odrl method}
\label{sec:learning_odrl}

\begin{figure}[t]
	\centering
	\includegraphics[width=1\textwidth]{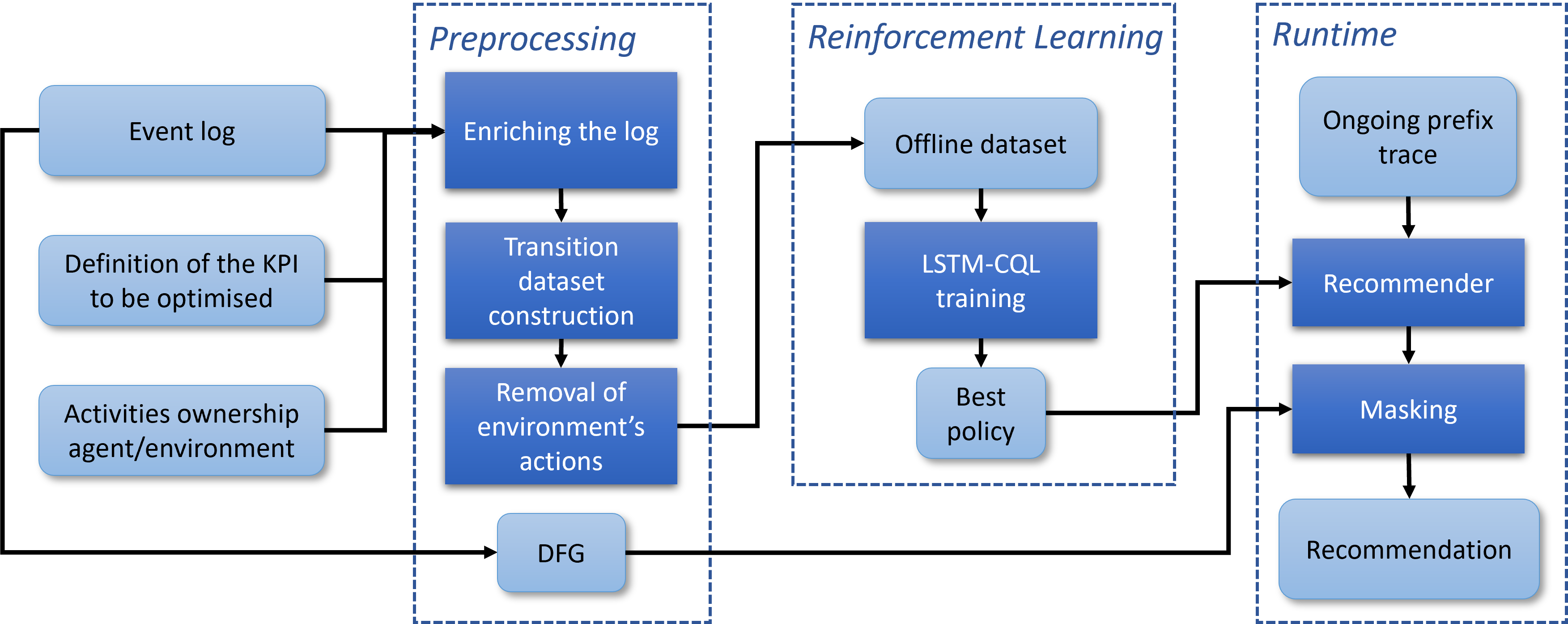} 
	\caption{Pipeline of the \odrl method.}  \label{fig:odrl_pipeline}
\end{figure}

The pipeline for the \odrl method is shown in Figure~\ref{fig:odrl_pipeline}.
This method is based on the adaptation of a state of the art technique in \odrl to the process mining domain, which learns a policy directly from event data without requiring interaction with the system.
Specifically, we adopt Conservative Q-Learning (CQL) \cite{CQL} (see Section~\ref{sec:background_RL}).
Similarly to other Deep RL models, the CQL algorithm uses neural networks to approximate the action-value function \eqref{eq:q_value}. Given the sequential nature of event log data, we instantiate a Recurrent Neural Network (RNN) encoder architecture, leveraging the Long Short Term Memory (LSTM) cells within the RNN network.
We adopt the DiscreteCQL implementation from the d3rlpy library\footnote{\url{https://d3rlpy.readthedocs.io/en/v2.2.0/references/algos.html\#discretecql}}
which is specialized in RL problems with discrete action spaces, as is the case in our scenario.
We customize this method by defining an LSTM encoder layer using the \emph{encoded factory} parameter.
This is a natural choice with sequential input and so it is suitable for our application,
and has already been adopted in a preliminary study using Deep Q-learning applied to PPM~\cite{10.1007/978-3-030-72693-5_10}.

\begin{figure}[t]
	\centering
	\includegraphics[width=1\textwidth]{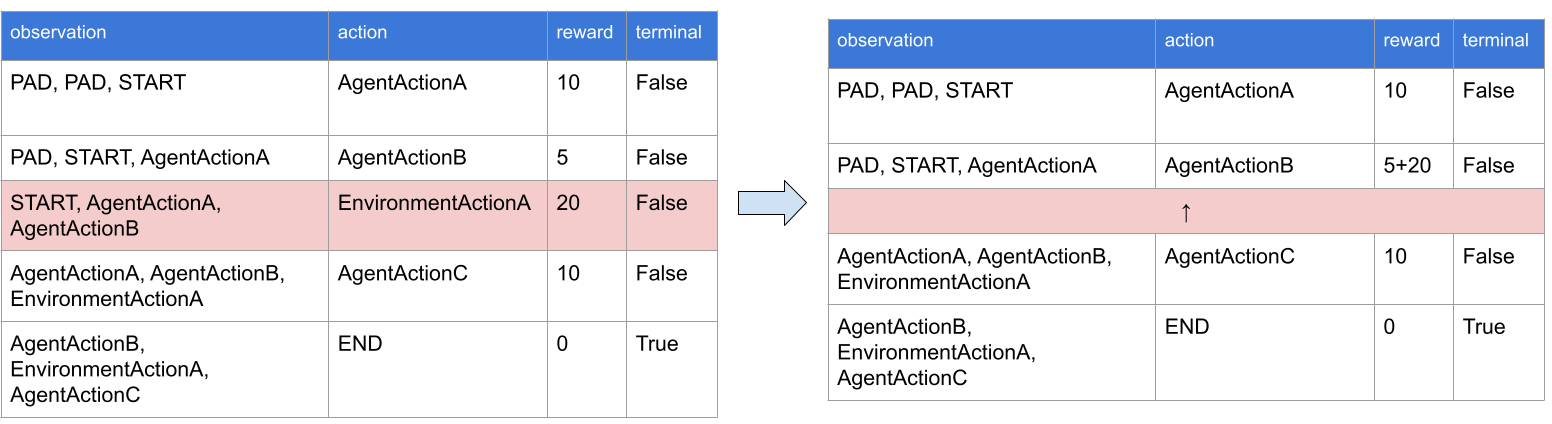}
	\caption{Removal example of a transition containing an environment action: the transition is removed from the dataset and its reward is added to the reward of the previous transition.}
    \label{fig:odrl_cumulate_rewards}
\end{figure}

In order to train the CQL agent, we need to extract a proper dataset for the relevant RL problem from the process log.
This dataset consists of a collection of transitions, each represented as a tuple with four elements: \emph{observation, action, reward, terminal}.
For every trace and every event in it, excluding the last one, the four elements of the transition are defined as follows:
\begin{itemize}
\item \emph{observation:} represents the current state, which is a vector encoding the current prefix (the trace truncated at the current event), properly encoded for input to the LSTM encoder layer. The encoding considers a window size with the most recent 20 events,\footnote{The window size was selected to be on the same order of magnitude as the average trace length of the considered logs reported in Table~\ref{tab:dataset}.} where padding is preprended in prefixes shorter than 20, and each activity is encoded using one-hot encoding.\footnote{This encoding, similar to the simple index encoding commonly used in PPM tasks~\cite{DBLP:conf/bpm/Francescomarino18}, captures only control-flow information, in line with the MDP-based method.}
\item \emph{action:} is the one-hot encoding of the activity of the next event in the trace.
\item \emph{reward:} is the reward obtained in the next event.\footnote{Unlike the MDP-based method, which considers the cumulative reward of the trace at the last event, in the preparation of this dataset we consider the instantaneous reward given by each action at each transition. We observed that this approach leads to faster convergence of the \odrl training.}
\item \emph{terminal:} is a boolean feature indicating if the current transition is the last of the trace.
\end{itemize}
At this point we have obtained a dataset which contains transition for all the actors in the process, to align this dataset with the RL problem under consideration, we need to remove all transitions where the \emph{action} corresponds to an activity associated with the environment.
In these cases, we transfer the reward of the deleted transition back to the previous valid transition, as shown in the toy example of Figure~\ref{fig:odrl_cumulate_rewards}. The edge case where the transition to be removed has the terminal flag to ``True'' is also handled by moving the flag to the previous transition.

As the primary goal of this work was to evaluate the two proposed RL methodologies, the CQL training parameters and the LSTM network parameters were chosen to ensure methodological consistency across experiments rather than to optimize performance for each dataset. Specifically, we adopted standard choices for both the LSTM architecture (hidden size 128, batch size 64) and the CQL training phase (number of epochs 100, steps per epoch 150,000, target update interval 32,000, $\gamma=1$), while keeping all other parameters at their default values.\footnote{\emph{number of steps per epoch} and \emph{target update interval} have been selected looking at the approximate mean number of transitions in the event logs, we selected $\gamma=1$ since our episodes are finite and positive reward is always given at the last activity.}


Finally, during the inference phase, we applied masking to the recommended actions by the agent. This masking was achieved by constructing a directly-follows graph from the training log using the pm4py library\footnote{\url{https://processintelligence.solutions/pm4py/}} and removing actions appearing in transitions that occurred less frequently than the 20th percentile.


\section{Evaluation setting}
\label{sec:evaluation}
In the following, we describe the evaluation setting designed to assess the performance of the policies learned with the two proposed RL methods.

\subsection{Research questions}
\label{sec:RQ}

The goal of the evaluation is to verify that the learned policies effectively recommend next activities that increase the value of the targeted KPI in a robust manner across different event log characteristics.
Moreover, we aim to analyze the differences between the two proposed methods (MDP-based RL and \odrl) in terms of policy performance and computational efficiency.
More in detail, we are interested to answer 
the following research questions:
\begin{enumerate}[label=\textbf{RQ\arabic*},leftmargin=2\parindent,topsep=1.5pt]
\item\label{RQ1} Which instantiations of the scaling function $h$ employed in the MDP-based method lead to the most robust policy performance in improving the targeted KPI across scenarios with varying event log sizes and success rates? 
\item\label{RQ2} How do the policies learned by the MDP-based method and by \odrl compare in terms of improving the targeted KPI across varying event log characteristics and in real-world scenarios?
\item\label{RQ3} How do the two RL methods: MDP-based and \odrl, compare in terms of computational efficiency?
\end{enumerate}
\ref{RQ1} aims to compare the different instantiations of the scaling function $h$ proposed in Section~\ref{sec:RLtraining}, and to evaluate whether they lead to more effective and robust policies than those generated using the standard choice $h^0 = 1$.
The aim of \ref{RQ2} is to compare the two proposed methods in terms of their ability to efficiently learn effective policies directly from event data across different scenarios, characterized by varying properties of the event logs.
Finally, \ref{RQ3} aims at evaluating the computational efficiency of the two proposed methods in learning the policies.

\subsection{Datasets description}
\label{sec:dataset}

To evaluate the performance of the proposed solution, we use two real-world, publicly available BPI Challenge event logs: BPIC2012 \cite{vandongen_2012} and BPIC2017 \cite{bpic17},\footnote{The BPIC2017 log contains two types of traces, identified by the value of the \emph{ApplicationType} attribute: either \emph{New credit} or \emph{Limit raise}. We filter the log to retain only the former, in order to focus on new loan applications.} as well as a set of eight synthetic event logs inspired by BPIC2012.
All these event logs describe the execution of a loan application process in a financial institution, which is qualitatively similar to the example presented in Section~\ref{sec:motivating_example}.
The process involves two actors: the bank issuing the loans and the customers applying for them. We assume the perspective of the bank and aim to learn a policy for selecting the next activity or the next activities, among those under its control, that increases the desired KPI.
Table~\ref{tab:env_activities} reports the set of activities not controlled by the bank for each event log considered; all remaining activities are assumed to be under the bank’s control and can therefore be selected by the policy.\footnote{We assume that the policy can select among all activities under the control of the bank (i.e., excluding those controlled by the customer). However, some of these activities may not be directly aligned with the optimization objective; this aspect is discussed in Section~\ref{sec:limitations}.}

\begin{table}
	\centering
	\resizebox{1\columnwidth}{!}{
	\begin{tabular}{ll}
			\toprule
			Datasets  & Environment activities  \\
			\midrule
            Synthetic / BPIC2012 & \actlabel{O\_Sent\_Back, O\_Declined, O\_Accepted, A\_Cancelled} \\
            BPIC2017 & \actlabel{O\_Returned, O\_Refused, O\_Accepted, A\_Cancelled  } \\
			\bottomrule
	\end{tabular}
	}
	\caption{Set of activities controlled by the environment for each event log type.}
	\label{tab:env_activities}
\end{table}

We define the KPI as the profit earned by the bank from issuing a loan, excluding the operational costs of executing the process. It consists of two components:
(i) a positive component, representing the profit gained when a loan offer is accepted by the customer—specifically, the interest on the loan, which we arbitrarily set at 15\% of the amount requested in the loan application; and
(ii) a negative component, representing the cost associated with the working time spent on the process\footnote{Working time is computed using the duration of the activities executed in the trace, where available. For activities lacking this information, a fixed arbitrary duration of 10 minutes is assumed.}, which we set at 36 euros per hour.

For the generation of the eight synthetic logs we define a simulation model $\mathcal{M}$, inspired by the BPIC2012 process.
Specifically, we extend the BPMN model in \figurename~\ref{fig:example} that represents the process model $\mathcal{N}$, and the process perspective parameters $\mathcal{P}$ are either estimated from the BPIC2012 event log or hypothesised when insufficient data is available.
In particular, the control flow parameter $\mathcal{P_C}$ is defined, for each gateway in the BPMN, as a set of conditional probabilities based on factors such as: the amount required in the loan application, the number of offers created and calls made by the bank during the process.\footnote{
The complete BPMN model and the $\mathcal{P}_C$ parameters are reported in~\ref{appx:simulation_2}, while the remaining simulation parameters used to define $\mathcal{M}$ are available in the evaluation repository.}
%
The main differences between the eight synthetic logs concern their sizes, that is the number of traces they contain,
and the probability that a loan application is (pre)accepted by the bank, which, in turn, affects the probability that a loan offer is accepted by a customer and hence the possibility for the bank to obtain a high KPI.

Table~\ref{tab:dataset} shows the details of the different event logs. Although the two real logs BPIC2012 and BPIC2017 represent the same overall process, there are several differences between them.
In particular the latter contains a larger number of traces, more variability among traces, longer traces on average and additional activities\footnote{
The BPIC2017 log contains also detailed information on the type of loan offer created by the bank, including number of terms, monthly cost, etc., which could
be used to obtain a finer policy. However, at this stage, we do not consider data payload in the definition of the RL problem.}
, making it a more complex test ground.
Furthermore, all the traces contained in BPIC2017 correspond to pre-accepted applications, for which at least one loan offer is created by the bank.
Synthetic logs are denoted as $\mathcal{L}_{size}$ and $\mathcal{L}_{size,rare}$
where $size$ is the number of traces contained in the log, and the presence of the $rare$ subscript indicates a low success rate of the log, that is a low percentage of traces achieving the acceptance of the offer by the customer (activity \actlabel{accept} as in \figurename~\ref{fig:example}), and hence a low average KPI value. This is achieved by imposing a low probability (less than 50\%) in the model to have a pre-accepted loan (nominally at the first gateway of BPMN model).
The differences between these eight synthetic logs will be central for answering \ref{RQ1} and \ref{RQ2}.


\begin{table}
	\centering
	\resizebox{1\columnwidth}{!}{
	\begin{tabular}{lcccccc}
			\toprule
			\multirow{2}*{Dataset}  & \multirow{2}*{Trace \#} & \multirow{2}*{Variant \#}  &  \multirow{2}*{Activity \#}  & Avg.trace  &	Application & Offers   \\
									&  						  &  						   &						   & length 	&   pre-accepted & accepted \\
			\midrule
			BPIC2012  						&  13087 	& 4366 			& 	24	& 20 			& 56\%  			&  17\% 			\\
			BPIC2017 (\emph{New credit})	&  28120	& 14120 		& 	26	& 39			& 100\%  			&  52\%  			\\
			$\mathcal{L}_{2000}$  			&  2000		& 509	& 	23	& 20 	& 59\% 	&  25\% 	\\
			$\mathcal{L}_{2000,rare}$ 		&  2000  	& 415	& 	23	& 16 	& 41\% 		&  18\% 	\\
			$\mathcal{L}_{4000}$  			&  4000 	& 880 	& 	23	& 20 	& 60\%  			&  25\% 	\\
			$\mathcal{L}_{4000,rare}$ 		&  4000  	& 610	& 	23	& 16 	& 40\% 				&  17\% 	\\
			$\mathcal{L}_{8000}$  			&  8000 	& 1454 	& 	23	& 21 	& 60\%  			&  27\% 	\\
			$\mathcal{L}_{8000,rare}$ 		&  8000  	& 1084 	& 	23	& 16 	& 40\% 				&  17\% 	\\
			$\mathcal{L}_{16000}$ 			&  16000 	& 2340 	& 	23	& 21 	& 60\%  			&  27\% 	\\
			$\mathcal{L}_{16000,rare}$ 		&  16000  	& 1778 	& 	23	& 16 	& 40\% 				&  18\% 	\\
			\bottomrule
	\end{tabular}
	}
	\caption{Datasets description.}
	\label{tab:dataset}
\end{table}

\subsection{Evaluation methods}
\label{sec:ev_methods}

\begin{figure}[ht]
  \centering
  \begin{subfigure}{0.65\textwidth}
    \centering
    \includegraphics[width=\linewidth]{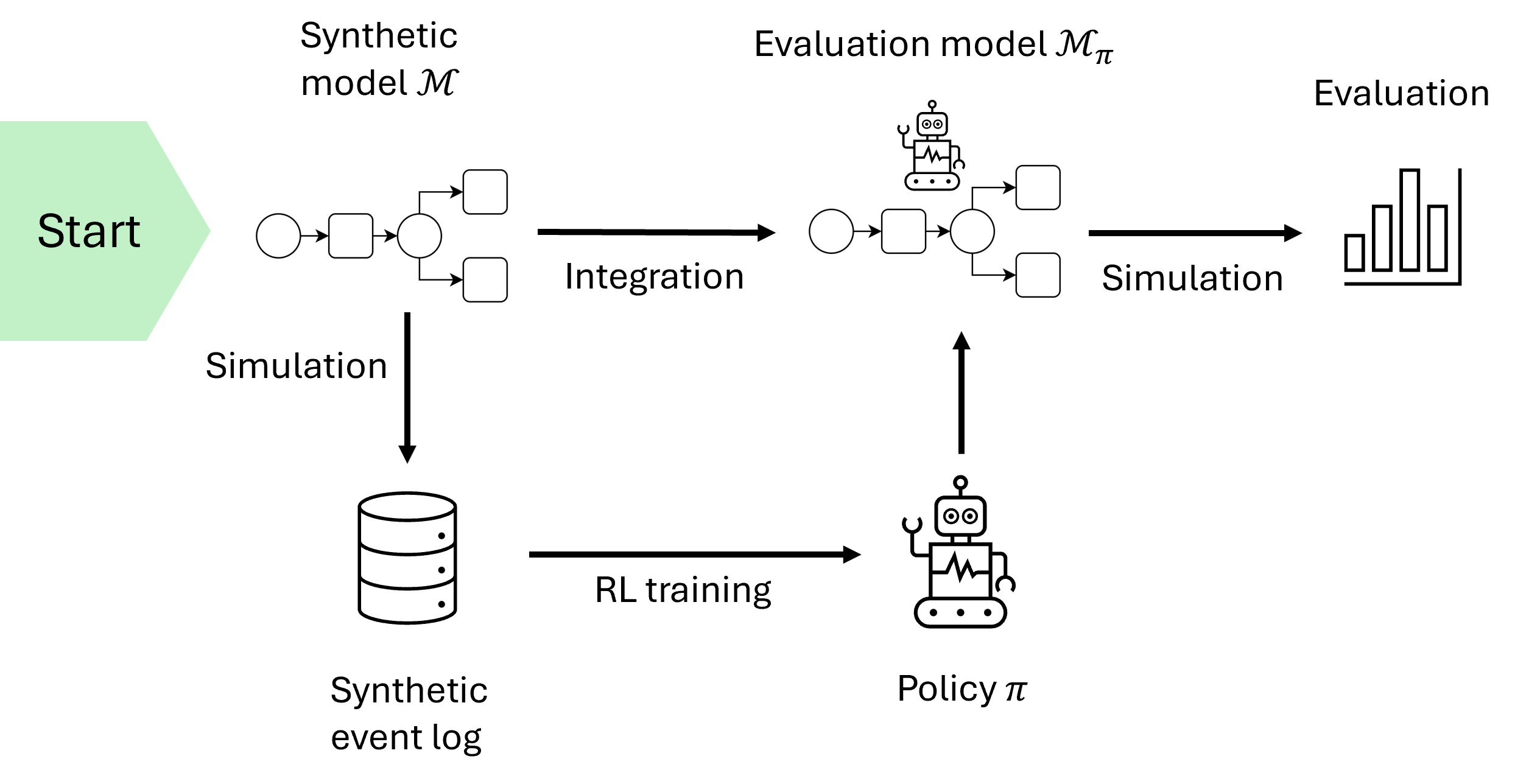}
    \caption{}
    \label{fig:syn_evaluation}
  \end{subfigure}
  
  \vspace{1em} 

  \begin{subfigure}{0.92\textwidth}
    \centering
    \includegraphics[width=\linewidth]{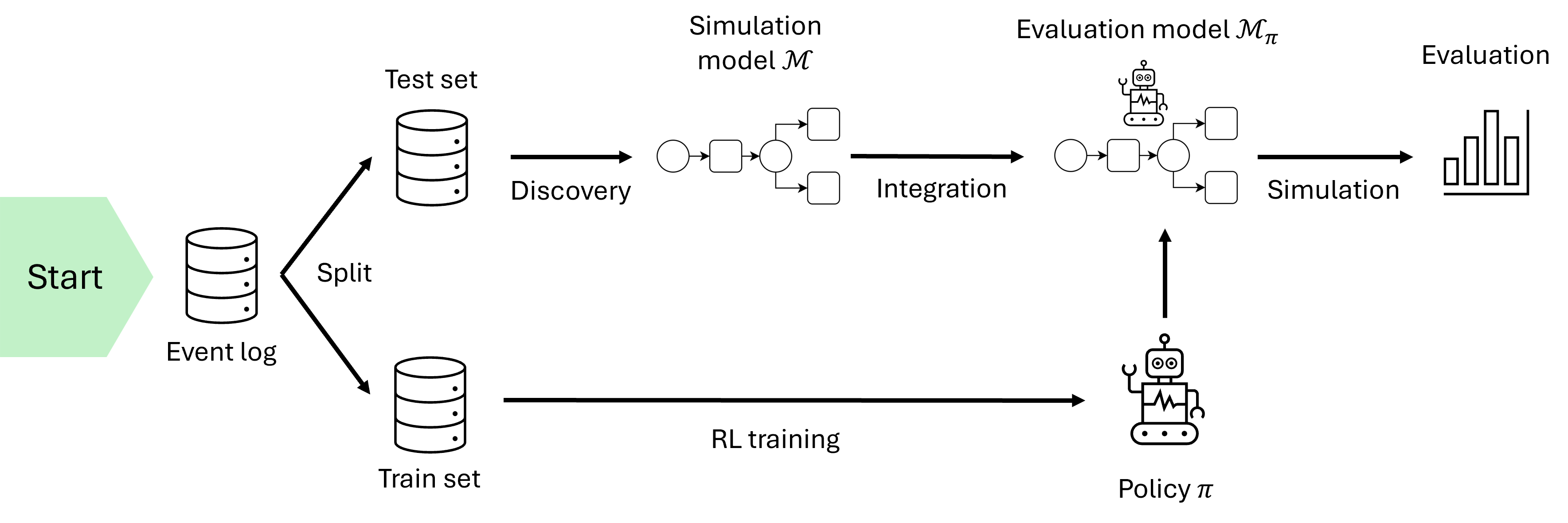}
    \caption{}
    \label{fig:real_evaluation}
  \end{subfigure}

  \caption{Evaluation pipeline for (a) the synthetic log and (b) the real-world log.}
  \label{fig:evaluation}
\end{figure}

A well-known problem in evaluating recommender systems for a real-world process scenario
is the difficulty of testing the recommendations provided~\cite{DBLP:conf/bpm/Dumas21}.
To circumvent this issue, we leverage Business Process Simulation (BPS) to evaluate the policy discovered on both synthetic and real logs, and to answer the research questions presented in Section~\ref{sec:RQ}.
\figurename~\ref{fig:evaluation} shows the evaluation pipeline for both the synthetic and real log cases.
Specifically, the simulation model used for evaluation is able to integrate the policy that recommends the next activity for the agent during execution. 
We therefore define $\mathcal{M}_\pi = (\mathcal{N}, \mathcal{P}, \pi)$, a modified simulation model of $\mathcal{M}$~(Section~\ref{sec:background_bps}), that takes a policy $\pi$ as input.
The policy controls the selection of those activities governed by the agent, but other activities can occur at any time, and they correspond to the environment's response to the agent's behavior.
Therefore, $\mathcal{M}_\pi$ is capable of simulating the interplay between the agent (the bank) and the environment (the customer).\footnote{Achieving this automatically through simulation, without any a priori knowledge of the process, is non-trivial and constitutes a novelty in the state of the art. To accurately reproduce the interactions between the environment and the agent, we leverage two Directly-Follows Graphs (DFGs): one is used to run the simulation, and the other to verify the compliance of the agent's possible actions. For instance, in Figure~\ref{fig:example}, at gateway $5$, the agent can respond with either \actlabel{ACCEPT} or \actlabel{CANCEL}. Both are valid, but only the former adheres to the BPMN model, while the latter implicitly assumes a skip in the simulation.}
After identifying which activities belong to the agent and which to the environment (see Figure~\ref{fig:example}), each simulation step in $\mathcal{N}$ is classified based on who is allowed to act. There are three possible types of steps: mixed steps, where either the agent or the environment can act; and either pure agent or pure environment steps, where only one of them is allowed to perform an activity.

In case of a pure environment step, the next activity is determined according to the semantics of $\mathcal{N}$ and $\mathcal{P}_C$; otherwise, in a pure agent step, $\pi$ is used to recommend the next activity.
In a mixed step, the environment has the first opportunity to act; if it refrains from doing so, the agent is then allowed to intervene. For example, at gateway $3$ in Figure~\ref{fig:example}, the environment may choose to act with activities \actlabel{Sent back offer}, \actlabel{Cancel} and \actlabel{Accept}. If it does not, the agent can proceed by recommending either \actlabel{Create offer} or \actlabel{Decline}.
Specifically, the policy requires as input the prefix $\sigma_k$ of the trace that is being simulated, and returns the recommended next activity.
The model $\mathcal{M}_\pi$ verifies if the recommended next activity is allowed, i.e., whether it complies with the BPMN model; 
if not, the trace simulation ends with an exception and the (negative) KPI computed at that point is considered\footnote{This case never occurs for any of the policies obtained in this work.}.
Finally, during the simulation, it may happen that a trace never reaches the end of the process because the policy keeps recommending activities that have negative effects on the environment response,
that is, it reduces to zero the probability of the environment to respond.
In this case, we stop the trace simulation after a maximum number of steps raising another exception, and we consider the (negative) KPI computed at that moment.

For both synthetic and real logs, we employ \rims~\cite{rims} as the BPS simulator, as it is a hybrid and flexible simulator that can be modified to incorporate a policy.


\paragraph{Synthetic logs evaluation}

The evaluation pipeline in the synthetic case is shown in \figurename~\ref{fig:syn_evaluation}.
The starting point is the simulation model $\mathcal{M}$ used for generating the synthetic log. $\mathcal{M}_\pi$ is obtained by integrating the policy $\pi$ learned from the data.
In particular, for every synthetic log $\{\mathcal{L}_{i}, \mathcal{L}_{i,rare}, i=2000,4000,8000,16000\}$ and RL method, we obtain a different policy $\pi$ which is used to define a specific evaluation model $\mathcal{M}_\pi$. 
Then, we use these models to simulate the behavior of the environment in response to the agent's recommendations over $2000$ traces and compute the average value of the obtained KPI, which is the metric used to evaluate and compare the techniques.


\paragraph{Real logs evaluation}

The evaluation pipeline in the real log case is shown in \figurename~\ref{fig:real_evaluation}.
In this case the starting point is the log itself, since no apriori simulation model is available.
The log is temporally split into a training set and a test set using an 80\%-20\% ratio.
The training set is used to train the RL agents and obtain the policy $\pi$, while the test set is used for discovering a simulation model $\mathcal{M}$ in which the policy is subsequently integrated, thus  obtaining an evaluation model $\mathcal{M}_\pi$.

To ensure a greater degree of freedom for policies in recommending the next activity, we leverage a ``spaghetti model'' obtained with the Inductive Miner algorithm~\cite{inductive} with a low threshold to filter out infrequent behavior, while the simulation parameters related to the BPMN model are defined using Simod~\cite{simod}.
In the proposed evaluation, the definition of $\mathcal{P}_C$ plays a fundamental role in governing the interaction between the environment and the agent. In particular, $\mathcal{P}_C$ refers to the decision point of $\mathcal{N}$. If this point corresponds to a pure environment step, the parameters determine which activity the environment performs. In the case of a mixed step, they also determine whether the environment chooses to act and, if so, which activities it performs. 

In BPS, the next activity at a decision point is typically determined using simple probabilities. However, in this way, the attributes of the ongoing trace and/or its history are not considered to define the likelihood of the next activity. As a result, the effect of policies may be undermined, since the sequence of chosen activities does not influence the environment’s decisions.

Therefore, we define two types of evaluation models $\mathcal{M}_\pi$ and $\mathcal{M}^\text{DT}_\pi$ that differ in how the environment determines its actions: the first uses fixed probabilities, while the second integrates Decision Trees into the simulation model, as described in~\cite{rims}.
Specifically, we train a Decision Tree for each decision point of $\mathcal{N}$, so that the environment action is determined based on the loan amount requested and the previously executed activities.
Finally, to properly evaluate the application of the policy in real-world processes, which exhibit more complex process behavior than synthetic logs, we perform $20$ simulations, each containing the same number of traces as the original test set used to discover the simulation model, and aggregate the results in terms of average KPI across them, ensuring sufficient statistical reliability.\footnote{
We consider $20$ simulations, as this is a common choice in the BPS literature~\cite{DBLP:conf/bpm/LopezPintadoRD25,DBLP:journals/softx/ChapelaCampaLSD25,DBLP:conf/icpm/KirchdorferBKAS24,SimBank}.}

\paragraph{Baselines}
We compare the proposed RL-based approaches against the following two non-learning reference policies:
\begin{itemize}
    \item For both the real and the synthetic cases, we consider as a baseline the stochastic policy embedded in the simulation model.
    In the real-log setting, this corresponds to the customary policy adopted in the real process and implicitly captured during the discovery of the simulation model.
    In the synthetic case, which is inspired by the real BPIC2012 process, this policy is explicitly defined to realistically reproduce the original process behavior, as detailed in Figure~\ref{fig:complete_BPMN_syn}.
    We refer to this policy as the \emph{customary policy} and denote it by $\pi_{\text{customary}}$.
    \item In addition, for the synthetic event logs, since we have access to the rules governing customer behavior (which are typically unknown in real-world settings), we define a strong heuristic policy that exploits this \emph{privileged knowledge} of the simulation logic to increase the probability of offer acceptance, as detailed in~\ref{appx:baseline_policy} and summarized in Table~\ref{tab:baseline_rules}.
    Importantly, this information is not directly available to the RL-based approaches, which must instead infer such behavioral patterns solely from observed executions.
    We refer to this as the \emph{informed heuristic policy} and denote it by $\pi_{\text{informed}}$.
\end{itemize}


\section{Evaluation results}
\label{sec:result}



In this section we report the results of the evaluation and we address the research questions of Section \ref{sec:RQ}. All the code and material are available at \url{https://github.com/rgraziosi-fbk/rl-optimal-policy-pm}.

\subsection{Answering RQ1}
\label{sec:answer_RQ1}

To answer~\ref{RQ1}, we compare different policies obtained through the MDP-based RL method described in Section~\ref{sec:learning_MDP}, using the various choices for the scaling function $h$ used in~\eqref{eq:scaled_q_value}, i.e., the step function from~\eqref{eq:hstep} and the sigmoid function from~\eqref{eq:hsig}.
Firstly we construct the MDP for all the dataset following the procedure described in Section~\ref{sec:constructionMDP}.\footnote{As explained in Section~\ref{sec:constructionMDP}, the number of cluster for each dataset is identified applying the elbow method to the Within-Cluster-Sum of Squared Errors (WSS). These numbers are 80 for $\mathcal{L}_{16000}$ and $\mathcal{L}_{8000,rare}$; 100 for BPIC2017, $\mathcal{L}_{2000}$, $\mathcal{L}_{4000}$, $\mathcal{L}_{8000}$,
$\mathcal{L}_{2000,rare}$,
$\mathcal{L}_{4000,rare}$, $\mathcal{L}_{16000,rare}$; and 140 for BPIC2012.}
Then we consider seven different instantiations of the scaling function during training: the trivial choice $h^0 = 1$, three parameter settings for the step function $h^\text{step}_{n_T}$ from~\eqref{eq:hstep}, with $n_T \in \{Q_1, Q_2, Q_3\}$, and three parameter settings for the sigmoid function $h^\text{sig}_{\lambda}$ (also defined in~\eqref{eq:hstep}), with $\lambda \in \{Q_1, Q_2, Q_3\}$.
The symbols $Q_1$, $Q_2$, and $Q_3$ correspond to dataset-dependent quantities, computed as the first, second, and third quartiles, respectively, of the number $n(s,a)$ of occurrences in the log of all state-action pairs appearing in the MDP. The values of these quantities for all the datasets are reported in Table~\ref{tab:occurencies_quartiles}. 
The seven policies derived are respectively denoted as $\pi^\text{MDP}_0$, $\pi^\text{MDP}_{\text{step}Q_1}$, $\pi^\text{MDP}_{\text{step}Q_2}$, $\pi^\text{MDP}_{\text{step}Q_3}$, $\pi^\text{MDP}_{\text{sig}Q_1}$, $\pi^\text{MDP}_{\text{sig}Q_2}$, $\pi^\text{MDP}_{\text{sig}Q_3}$.

\begin{table}[t]
\centering
\begin{tabular}{lccc}
\toprule
Dataset & $Q_1$ & $Q_2$ & $Q_3$ \\
\midrule
BPIC2012                        & 2 & 11 & 55     	\\
BPIC2017 (\emph{New credit})    & 2 & 8  & 47   	\\
$\mathcal{L}_{2000}$            & 2 & 10 & 41 	\\
$\mathcal{L}_{4000}$ 			& 2 & 9  & 56 	\\
$\mathcal{L}_{8000}$            & 3 & 15 & 106  \\
$\mathcal{L}_{16000}$ 			& 3 & 14 & 161  \\
$\mathcal{L}_{2000,rare}$       & 2 & 9  & 33   \\
$\mathcal{L}_{4000,rare}$ 		& 2 & 9  & 52   \\
$\mathcal{L}_{8000,rare}$       & 2 & 13 & 90   \\
$\mathcal{L}_{16000,rare}$ 		& 3 & 12 & 115 	\\
\bottomrule
\end{tabular}
\caption{Quartiles ($Q_1$, $Q_2$, $Q_3$) of the number of occurrences $n(s,a)$ of state-action pairs in the log for each dataset.}
\label{tab:occurencies_quartiles}
\end{table}

To assess the robustness of each policy’s performance, we apply the evaluation methodology from Section~\ref{sec:ev_methods} to the policies derived using different instantiations of the scaling function $h$. We train them on the eight synthetic logs listed in Table~\ref{tab:dataset}, which cover a range of characteristics of the event logs used to train the policies.

\begin{figure}[t]
	\centering
	\includegraphics[width=\textwidth]{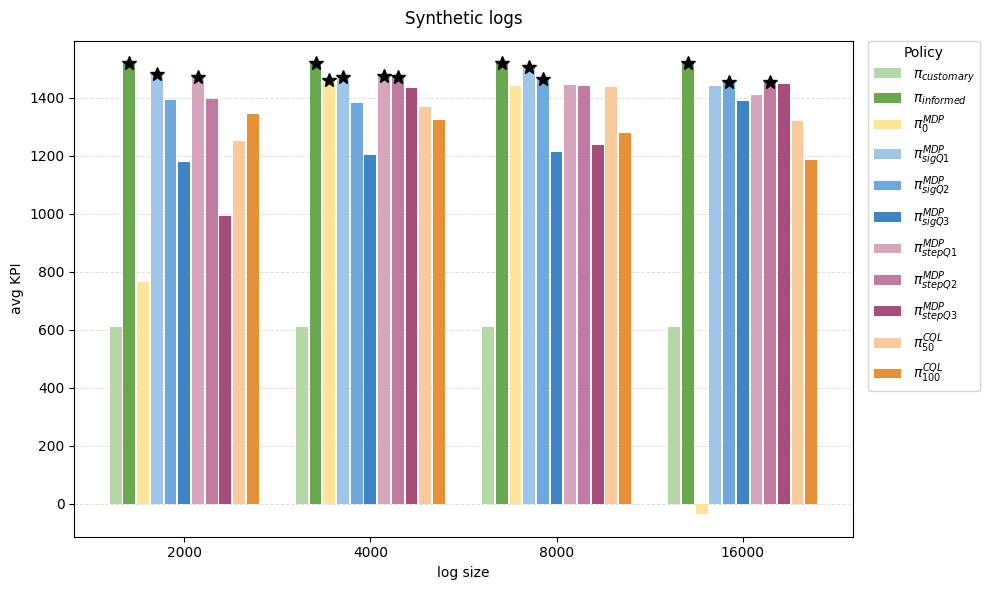} 
	\caption{Evaluation of the MDP policies in the synthetic logs $\mathcal{L}_i$.}
        \label{fig:MDP_syn_log}
\end{figure}

\begin{figure}[t]
	\centering
	\includegraphics[width=\textwidth]{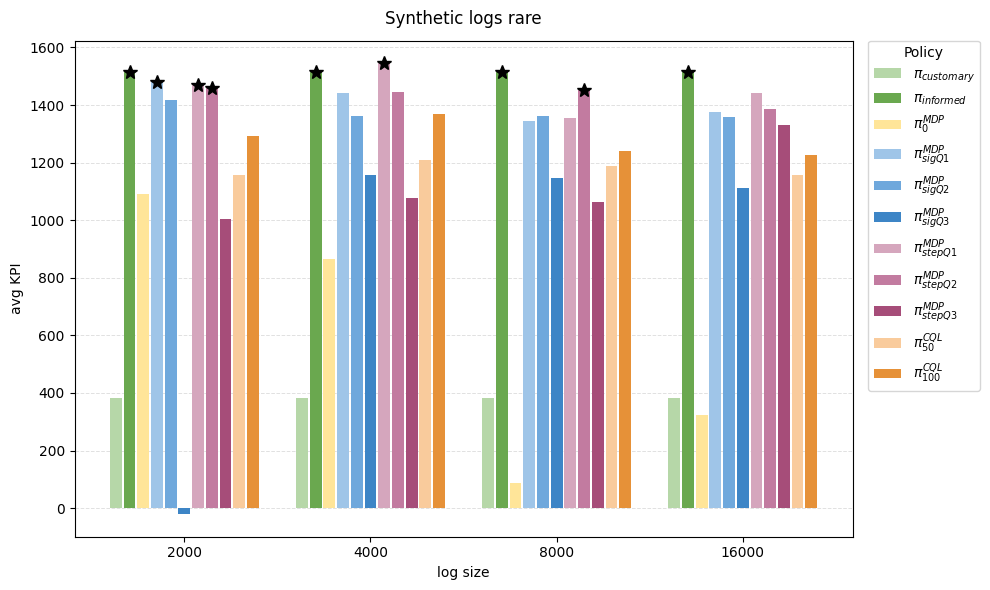} 
	\caption{Evaluation of the MDP policies in the synthetic logs $\mathcal{L}_{i,rare}$.}
        \label{fig:MDP_syn_log_rare}
\end{figure}

Figure~\ref{fig:MDP_syn_log} and Figure~\ref{fig:MDP_syn_log_rare} report the average KPI obtained on the synthetic $\mathcal{L}_i$ and $\mathcal{L}_{i,rare}$ logs, respectively.
For each log, the policy achieving the highest average KPI, as well as those whose performance is not statistically significantly different from the best, are marked with a black star above the corresponding bar.\footnote{Statistical significance is assessed using a two-sample t-test on the KPI values of the simulated traces. Differences are considered statistically significant when the $p$-value is below $0.05$.}


The first and second columns in each plot correspond to the average KPI achieved using the \emph{customary policy} integrated into the simulation model $\mathcal{M}$ and the \emph{informed heuristic policy}, respectively, as described at the end of Section~\ref{sec:ev_methods}. Since these policies are not learned, their performance is independent of the training log size; therefore, their values are reported identically across all sizes to facilitate visual comparison.

%
Preliminarily, we observe that the customary policy---i.e., the policy underlying the synthetic data used for training---is almost always outperformed by the learned policies.
Second, we observe that the informed policy always belongs to the best-performing policies across all synthetic logs.
This is expected, as the heuristic defining it relies on perfect knowledge of customer behavior (see~\ref{appx:baseline_policy}), which would typically be unavailable a priori,
while the learned policies must instead be inferred offline from data.
However, we observe that in all datasets, with the single exception of $\mathcal{L}_{16000,\text{rare}}$, there is at least one learned policy whose performance is statistically indistinguishable from that of the informed heuristic.
This indicates that, with few exceptions (analyzed later), the proposed method can effectively learn policies that significantly improve the KPI based solely on historical executions.

Comparing the different choices of the scaling function $h$, we first observe that the naive choice $h^0 = 1$ (third column in each plot) leads to highly unstable performance across datasets, sometimes even worse than the customary policy implemented in the simulator ($\mathcal{L}_{16000}$, $\mathcal{L}_{8000,rare}$, $\mathcal{L}_{16000,rare}$). 
We attribute this poor and unstable behavior to the RL agent exploiting transitions captured in the MDP that appear promising but were observed only infrequently in the event log, and are therefore unreliable for learning. As explained in Section~\ref{sec:RLtraining}, the introduction of non-trivial scaling functions $h$ is intended precisely to mitigate this issue.

When considering the policies obtained with non-trivial choices of the scaling function, we generally observe a consistent improvement over the naive choice $h^0$, with only a single exception given by the $h^{\text{sig}}_{Q_3}$ choice for the $\mathcal{L}_{2000,rare}$ log, where it even gets negative average KPI, indicating very poor performance. 
More generally, we can observe that both $h^{\text{sig}}_{Q_3}$ and $h^{\text{step}}_{Q_3}$ seem to exhibit less stable performance compared to the other choices of the scaling function.
This behavior is likely due to the fact that both scaling functions penalize a very large set of transitions, including relatively frequent ones, and therefore end up trusting only the most frequent transitions in the MDP, 
thus becoming more sensitive to the specific state representation induced by the clustering step used to construct the MDP, an issue to which more robust scaling functions are less prone.
This issue is particularly pronounced when the size of the log is small, as suggested by the poorer performance of the two policies on the smallest logs compared to the larger ones.
Specifically, in the case of the $\mathcal{L}_{2000,rare}$ log, this can bias the learned policy toward overly conservative strategies, such as systematically declining applications, thus causing the occasional catastrophic performance of $h^{\text{sig}}_{Q_3}$.
We discuss this further at the end of the Section, where we provide a qualitative analysis of some of the policies learned.

All the other choices of the scaling function exhibit much more stable behavior across all datasets. This suggests that incorporating an appropriate scaling function $h$ in \eqref{eq:scaled_q_value} can enhance policy performance by discouraging misleading decisions based on underrepresented transitions in the training log.
In particular, these configurations consistently achieve performance comparable to, and sometimes equal to, the informed heuristic. This indicates that the learned policies are able to capture customer behavior effectively, despite not having any a-priori knowledge about it.

From the analysis of the results, the most robust choices for the scaling function are $h^\text{sig}_{Q_1}$, $h^\text{step}_{Q_1}$ and $h^\text{step}_{Q_2}$, as they more often produce policies that belong to the set of best-performing ones, namely in 4 out of 8 logs.
This behavior suggests that the primary role of the scaling function is effectively achieved when it attenuates the impact of low-frequency transitions in the decision process, largely independently of the specific functional form, as long as it does not excessively impact the agent’s exploration. In this regard, choosing the first quartile $Q_1$ or the median $Q_2$ as the threshold parameter for the scaling functions results in a moderate regularization effect: transitions supported by limited evidence are sufficiently down-weighted, while those backed by a reasonable number of observations retain a significant impact on the learned value function.
This balance reduces overfitting to spurious or noisy transitions while preserving informative structure in the MDP, thereby improving policy stability and generalization across different datasets.

\paragraph{Qualitative analysis of learned policies}
We provide here a qualitative analysis of selected policies learned from the synthetic logs, aimed at interpreting their performance in light of the complete knowledge of the simulation dynamics described in~\ref{appx:simulation_2}.

In particular, we focus on the $\mathcal{L}_{2000,rare}$ log, which is among the most challenging ones, and we analyze the informed heuristic $\pi_\text{informed}$, and three learned policies, namely $\pi^{\text{MDP}}_{\text{sig}Q_1}$, $\pi^{\text{MDP}}_{0}$, and $\pi^{\text{MDP}}_{\text{sig}Q_3}$. These correspond, respectively, to one of the best-performing policies, one with comparatively lower performance, and one that completely fails the learning task achieving a negative average KPI.

Table~\ref{tab:average_count_all_trace} reports the percentage of trace outcomes obtained under each policy, together with the average trace length.
We distinguish four types of outcomes: \emph{accepted}, where the loan is accepted by the customer; \emph{rejected by customer}, where the customer either declines the loan offer or cancels the application; \emph{rejected by bank}, where the bank decides to decline the application;\footnote{Specifically, traces are labeled as \emph{accepted} if they contain the activity \actlabel{O\_ACCEPTED}; \emph{rejected by customer} if they contain either \actlabel{O\_DECLINED} or \actlabel{A\_CANCELLED}; and \emph{rejected by bank} if they contain the activity \actlabel{A\_DECLINED} but not \actlabel{O\_DECLINED}.} and \emph{error}, where the policy produces an error because it reaches the maximum number of steps without reaching a valid trace ending.

We observe that the best-performing policy, $\pi^{\text{MDP}}_{\text{sig}Q_1}$, achieves the highest rate of accepted traces, very close to the rate obtained by the informed heuristic.
The compatible performance in average KPI with respect to the informed heuristic $\pi_\text{informed}$ is explained by a very close acceptance rate, together with a similar average trace length.
%
%
The lower performance of the $\pi^{\text{MDP}}_{0}$ policy is associated with an acceptance rate that is 16 percentage points lower, indicating that this policy is less effective in achieving acceptance from the customer.
Also the behavior of the worst-performing policy, $\pi^{\text{MDP}}_{\text{sig}Q_3}$, is clear. This policy has effectively learned to reject almost all incoming applications. As discussed earlier, this is likely due to the high reference length used in the sigmoid scaling, which strongly penalizes longer executions, even when they consist of relatively frequent and meaningful transitions, thereby biasing the policy toward prematurely terminating traces.
Finally, we observe that the best-performing policy $\pi^{\text{MDP}}_{\text{sig}Q_1}$ has learned to accept all applications in order to maximize the likelihood of assigning loans. This behavior is consistent with the KPI definition and the available information, which do not include long-term factors such as repayment, risk, or resource limitations in processing applications. As a result, rejection decisions are not explicitly rewarded, while only the costs associated with activity execution are taken into account. The importance of the KPI definition for the decision problem is further discussed in Section~\ref{sec:limitations}.

Table~\ref{tab:average_count_ACCEPTED_trace} reports the average and maximum frequencies of the key activities that influence customer behavior (see~\ref{appx:simulation_2}), computed only over traces with an \emph{accepted} outcome.
We exclude the $\pi^{\text{MDP}}_{\text{sig}Q_3}$ policy from this analysis, as it does not produce any accepted traces.
The values observed for the informed heuristic $\pi_\text{informed}$ are consistent with its rule-based definition (see~\ref{appx:baseline_policy}).
The best-performing learned policy exhibits a similar overall behavior, in particular by keeping the number of active offers below three, but it also shows notable differences
that reflect a more permissive approach toward activities that have no, or only slightly negative, influence on offer acceptance.
Specifically, the maximum frequencies of the activities \actlabel{W\_Call\_after\_offer} and \actlabel{W\_Call\_missing\_information} are higher than in the heuristic.
This reflects the exploratory nature of the RL algorithm, which may retain actions with marginal or uncertain impact when they are not sufficiently penalized by the reward structure.
Conversely, we observe a more cautious use of the \actlabel{O\_CREATED} activity by the learned policy, which never reaches three created offers. While this could, in principle, limit opportunities for customer acceptance, we observe that the average number of created offers remains very close to that of the informed baseline, indicating that the policy maintains comparable acceptance effectiveness while adopting a more conservative offer generation strategy, avoiding overly risky decisions, as increasing the number of offers is beneficial only up to three, beyond which acceptance decreases (see Table~\ref{tab:o_sent_back_prob}).

\begin{table}[h!]
\centering
\begin{tabular}{l|cccc}
\toprule
 \multirow{ 2}{*}{Traces}& \multicolumn{4}{c}{Policy} \\
 &$\pi_{\text{informed}}$ & $\pi^\text{MDP}_0$ & $\pi^\text{MDP}_{\text{sig}Q_1}$ & $\pi^\text{MDP}_{\text{sig}Q_3}$ \\
\midrule
\% \emph{Accepted} & 74.3\% & 58.4\% & 74.4\% & 0 \\
\% \emph{Rejected by bank} &1.75\% & 0 & 0 & 90\% \\
\% \emph{Rejected by customer}  & 23.95\% & 41.6\% & 25.6\% & 10\% \\
\% Error  & 0 & 0 & 0 & 0 \\
\midrule
Avg. Length  & 18 & 21 & 20 & 3 \\
\bottomrule
\end{tabular}
\caption{Distribution of trace outcomes type and average trace length under different policies for the synthetic log $\mathcal{L}_{2000,\text{rare}}$.}
\label{tab:average_count_all_trace}
\end{table}

\begin{table}[ht]
\centering
\begin{tabular}{l|cc|cc|cc} 
\toprule
 \multirow{ 2}{*}{Activity} 
& \multicolumn{2}{c}{$\pi_{\text{informed}}$} 
& \multicolumn{2}{c}{$\pi^\text{MDP}_0$} 
& \multicolumn{2}{c}{$\pi^\text{MDP}_{\text{sig}Q_1}$} \\
\cmidrule(lr){2-3} \cmidrule(lr){4-5} \cmidrule(lr){6-7} 
& avg & max & avg & max & avg & max \\
\midrule
Active offers
 & 1.2 & 2 & 3.1 & 9 & 1.0 & 2 \\
\actlabel{W\_Call\_after\_offer}
 & 1.6 & 7 & 2.9 & 13 & 1.6 & 9 \\
\actlabel{W\_Call\_missing\_information} 
 & 0 & 0 & 0.1 & 4 & 1.7 & 5 \\
\actlabel{O\_CREATED} 
 & 2.0 & 3 & 3.1 & 6 & 1.9 & 2 \\
\bottomrule
\end{tabular}
\caption{Average and maximum frequencies of key activities affecting customer behavior, computed over \emph{accepted} traces only.}
\label{tab:average_count_ACCEPTED_trace}
\end{table}

\subsection{Answering RQ2}
\label{sec:answer_RQ2}

In order to answer \textbf{RQ2}, we compare the performance of the most robust policies identified in \textbf{RQ1}, namely $\pi^\text{MDP}_{\text{sig}Q_1}$, $\pi^\text{MDP}_{\text{step}Q_1}$ and $\pi^\text{MDP}_{\text{step}Q_2}$, with the proposed \odrl method based on CQL, across all logs listed in Table~\ref{tab:dataset}.
In particular, we evaluate how the two approaches are able to learn effective policies under varying dataset sizes, success rates, and real-world scenarios.
Since no specific early stopping mechanism is applied during CQL training, we consider two policies obtained with this method, namely $\pi^\text{CQL}_{50}$ and $\pi^\text{CQL}_{100}$, corresponding to the models after 50 and 100 training epochs, respectively.\footnote{We verified convergence of the learning process by observing the stabilization of the estimated average value function over training.}

The results for answering RQ2 are reported in Figures~\ref{fig:MDP_syn_log} and~\ref{fig:MDP_syn_log_rare} for the eight synthetic logs, and in Table~\ref{tab:real_logs_results} for the real-world event logs.

We first compare the performance of the two approaches on the eight synthetic logs shown in Figures~\ref{fig:MDP_syn_log} and~\ref{fig:MDP_syn_log_rare}.
First, we observe that both approaches learn highly effective policies. In particular, for the $\mathcal{L}_i$ training logs, they are able to more than double the average KPI achieved by the customary policy, while for the $\mathcal{L}_{i,rare}$ training logs they more than triple it in most cases.



However, we observe a clear difference between the performance of the MDP-based policies and those obtained through \odrl. Indeed, the latter class never appears among the best-performing policies across all synthetic datasets, while each of the best-selected MDP-based policies appears among the best-performing ones in 4 out of 8 event logs.
This suggests that the MDP-based policies, with a proper choice of the scaling function $h$, are able to learn effective next-activity policies that generalize well across different event log characteristics, although the performance gap with the CQL-based approach remains moderate.

Finally, we observe no notable correlation between the performance of the two approaches and the log size, indicating that both methods are robust in learning good policies even from small datasets or when the representativeness of good behavior is scarce, as in the $\mathcal{L}_{2000,rare}$ log.
An additional analysis is shown in~\ref{appx:simulation}, where policies are applied from a fixed point in the trace, which confirms the findings of this section.

Concerning the real logs, Table~\ref{tab:real_logs_results} reports the average KPI obtained over 20 simulations\footnote{Each simulation contains the same amount of traces of the corresponding original test set, namely $2617$ and $5623$ for BPIC2012 and BPIC2017, respectively.} for each policy, using both types of simulation models: $\mathcal{M}_\pi$, which relies on fixed probabilities at decision points, and $\mathcal{M}_\pi^{\text{DT}}$, which instead employs decision trees.

We compare the learned policies against the customary policy (representing the stochastic policy adopted in the original real-world process) simulated on both models, $\mathcal{M}_\pi$ and $\mathcal{M}_\pi^{\text{DT}}$.
This approach is preferable to a direct comparison with the average KPI computed from the real logs, as it allows us to assess the improvements introduced by the policies while avoiding the noise and inaccuracies that may arise from the discovery of simulation models from the logs. Nevertheless, the average KPI obtained from simulations using the customary policy remains close to the values observed in the corresponding test event logs used to discover the models: $328$ for BPIC2012 and $1299$ for BPIC2017.

For each simulation model, we performed pairwise statistical significance tests. Underlined values in Table~\ref{tab:real_logs_results} indicate a statistically significant improvement over the customary policy, while boldface values indicate no statistically significant difference from the best-performing policy for that model.\footnote{Statistical significance is assessed using a two-sample t-test on the KPI values of the simulated traces. Differences are considered statistically significant when the $p$-value is below $0.05$.}

Firstly, we observe that all the learned policies outperform the customary policy, achieving a higher average KPI across all models. This demonstrates the effectiveness of both approaches in learning good policies also in real-world scenarios.

When comparing the different learned policies, we observe that under the model $\mathcal{M}_{\pi}$, almost all policies achieve statistically indistinguishable performance.
When using $\mathcal{M}^{\text{DT}}_{\pi}$, the performance of the policies becomes more differentiated, with $\pi^\text{MDP}_{\text{step}{Q_1}}$ emerging as the sole best-performing policy in the BPIC2012 log, and $\pi^\text{MDP}_{\text{sig}{Q_1}}$ jointly with the two CQL policies in the BPIC2017 log.

We attribute this difference to the higher sensitivity of the $\mathcal{M}^{\text{DT}}_{\pi}$ model. By leveraging decision trees to model gateway probabilities, this model provides an environment representation that is more sensitive to the history of the process instance. This enables a more refined assessment of the impact of different policies, compared to the $\mathcal{M}_{\pi}$ model, where gateway behavior is governed by fixed probabilities, limiting the influence that policies can exert on the environment’s response.

Overall, the comparison between the MDP-based and the CQL-based policies reveals solid performance for both approaches, with the MDP-based method showing a consistent advantage in the synthetic evaluation and performing slightly better or comparably to the \odrl method on the real logs. Therefore, the results indicate a consistent, though not overwhelming, advantage of the MDP-based approach across the different settings.

\begin{table}[h]
\centering
\small
\renewcommand{\arraystretch}{1.25} 
\setlength{\tabcolsep}{4pt}
\begin{tabular}{llcc}
\toprule
Log & Policy & \multicolumn{2}{c}{Average KPI} \\
& & Simulation Model $\mathcal{M}_\pi$ & Simulation Model $\mathcal{M}_\pi^{\text{DT}}$ \\
\midrule
\multirow{5}{*}{\rotatebox[origin=c]{90}{BPIC2012}} 
  & customary policy        & 274.780 & 688.403 \\
    & $\pi^\text{MDP}_{\text{sig}Q_1}$ & \underline{\textbf{374.966}} & \underline{886.098} \\
  & $\pi^\text{MDP}_{\text{step}Q_1}$     & \underline{\textbf{368.507}} & \underline{\textbf{914.636}} \\
  & $\pi^\text{MDP}_{\text{step}Q_2}$ & \textbf{\underline{366.831}} & \underline{864.043}  \\
  & $\pi^\text{CQL}_{50}$  & \underline{\textbf{369.776}} & \underline{851.857} \\
  & $\pi^\text{CQL}_{100}$   & \underline{\textbf{373.028}} & \underline{854.281} \\
\midrule
\multirow{5}{*}{\rotatebox[origin=c]{90}{BPIC2017}} 
  & customary policy        & 1598.043 & 1135.541 \\
    & $\pi^\text{MDP}_{\text{sig}Q_1}$   & \underline{\textbf{1731.108}} & \underline{\textbf{1487.583}} \\
  & $\pi^\text{MDP}_{\text{step}Q_1}$      & \underline{\textbf{1713.294}} & \underline{1420.352} \\
    & $\pi^\text{MDP}_{\text{step}Q_2}$      & \underline{\textbf{1709.773}} & \underline{1426.082} \\
  & $\pi^\text{CQL}_{50}$  & \underline{\textbf{1714.804}} & \underline{\textbf{1469.534}} \\
  & $\pi^\text{CQL}_{100}$   & \underline{\textbf{1730.305}} & \underline{\textbf{1495.888}} \\
\bottomrule
\end{tabular}
\caption{Comparison of policies on the BPIC12 and BPIC17 real logs, based on the average KPI computed over 20 simulations for each policy using both simulation models, $\mathcal{M}_\pi$ and $\mathcal{M}_\pi^{\text{DT}}$.}
\label{tab:real_logs_results}
\end{table}


\subsection{Answering RQ3}
\label{sec:answer_RQ3}

Finally, we analyze the computational efficiency of the two methods proposed in this paper for learning the best activity policies in order to answer to \ref{RQ3}.
In Table~\ref{tab:computation_time}, we report the time required to obtain the policy using both methods for all datasets considered in the evaluation. Specifically, the time reported for the MDP-based method refers to the total time required by the pipeline described in Section~\ref{sec:learning_MDP}, including the preprocessing step, the construction of the MDP, and policy learning via the Dynamic Programming algorithm.\footnote{In Table~\ref{tab:computation_time}, we report the maximum time required among all the policies described in Section~\ref{sec:answer_RQ1}.}
For the \odrl method, the table reports the training time per epoch, together with the total training time for the two configurations considered in our evaluation, i.e., after 50 and 100 epochs, respectively.

\begin{table}[ht]
\centering
\small
\renewcommand{\arraystretch}{1.25} 
\setlength{\tabcolsep}{4pt}
\begin{tabular}{lcccc}
\toprule
 \multirow{ 2}{*}{Dataset} & \multirow{ 2}{*}{MDP-based} & \multicolumn{3}{c}{Offline DRL} \\
\cline{3-5}
 &  & per epoch & Total (50 epochs) & Total (100 epochs) \\
\midrule
BPIC2012 						& 91 			& 1670 & 83500 & 167000 \\
BPIC2017 (\emph{New credit}) 	& 364 			& 1751 	& 87550 & 175100 \\
$\mathcal{L}_{2000}$ 			& 37 	& 1865 & 93250 & 186500 \\
$\mathcal{L}_{4000}$ 			& 56 	& 1909 & 95450 & 190900 \\
$\mathcal{L}_{8000}$ 			& 111 	& 1845 & 92250 & 184500 \\
$\mathcal{L}_{16000}$			& 209 	& 1858 & 92900 & 185800 \\
$\mathcal{L}_{2000,rare}$  		& 30 	& 1890 & 94500 & 189000 \\
$\mathcal{L}_{4000,rare}$ 		& 55 	& 1918 & 95900 & 191800 \\
$\mathcal{L}_{8000,rare}$  		& 92 	& 1924 & 96200 & 192400 \\
$\mathcal{L}_{16000,rare}$ 		& 171 	& 1812 & 90600 & 181200 \\
\bottomrule
\end{tabular}
\caption{Computation times (in seconds) for the MDP-based and Offline DRL methods across datasets.}
\label{tab:computation_time}
\end{table}

We can immediately observe that the time required by the MDP-based method is significantly lower compared to the \odrl method, especially 
compared to the total time required to train and effective policy.
This is somewhat expected, as deep learning models generally involve a higher computational cost for training.

We can also notice that the time required per epoch by the \odrl method remains relatively constant across datasets, around 30 minutes per epoch, indicating that its computational cost does not strongly depend on the log size. In contrast, the time required by the MDP-based method increases with the size of the event log. This growth is primarily due to the clustering phase and the construction of the MDP, both of which scale with the amount of data. On the other hand, the Dynamic Programming algorithm used for policy learning remains highly efficient, consistently requiring only a few seconds.

In conclusion we can claim that the MDP-based method is more efficient than the \odrl one.

\subsection{Discussion}
\label{sec:discussion}

The results of our evaluation provide an insightful comparison between MDP-based and \odrl methods for learning activity-level policies directly from event data. We first showed that an appropriate choice of the scaling function $h$ enables the MDP-based RL approach to learn policies that robustly improve the targeted KPI across scenarios characterized by different properties of the training event logs.
%
Secondly, in the comparison with the \odrl approach based on CQL, we observed that both methods achieve solid performance across all analyzed scenarios, with a consistent advantage of the MDP-based approach in both the synthetic evaluation and the real-world event log.

In terms of computational efficiency, the MDP-based approach is significantly more efficient, requiring only a fraction of the time needed by CQL to produce a viable policy. Overall, these findings suggest that the MDP-based method offers a favorable balance between simplicity, interpretability, efficiency, and performance. One possible explanation is the relatively lower complexity of event data compared to domains where the flexibility of deep learning models becomes essential. In process mining scenarios, the structured nature of event logs and the presence of well-defined decision points may already provide a sufficient foundation for learning effective policies using simpler, more interpretable models.

Overall, the results show that learning intervention policies directly from event data, under realistic assumptions of partial controllability, can effectively support KPI-oriented decision-making. The consistent performance across diverse logs and the computational efficiency of the MDP-based approach highlight the practical potential of the proposed framework for real-world process environments.


\section{Limitations}
\label{sec:limitations}

\paragraph{Generality of the Evaluated Scenario}
The main limitation of this study is the focus on a specific business process for defining our RL problem, namely the one described in Section~\ref{sec:motivating_example}. 
We mitigate this limitation by evaluating different types of event logs related to this problem, including both real-world and synthetic logs, and by considering various statistical characteristics of the event logs used to learn the policies.

\paragraph{Design Choices}
The objective of this work was to compare the effectiveness of two fundamentally different RL paradigms in learning behavioral policies directly from event log data. Instantiating these approaches for the targeted problem required several design choices, for which we provided reasoned justifications rather than conducting an exhaustive hyperparameter or architectural search, as such an exploration would have significantly increased the complexity and length of the paper, potentially harming its readability.
Nevertheless, we argue that the adequacy of these choices is supported a posteriori by the empirical results. In particular, once the key design components are properly configured (e.g., the scaling function in the MDP-based approach), both methods demonstrate strong, consistent performance across a diverse set of logs with varying statistical characteristics. This suggests that the proposed approaches are not overly sensitive to specific design decisions and that the main conclusions of the study do not rely on narrowly tuned configurations.


\paragraph{State Representation}
Another possible limitation of our study is the focus on the control-flow perspective, which serves as the main source of information for defining the state space in both the proposed methods. While the data payload is indirectly taken into account through its influence on the reward function, it is not explicitly modeled in the state representation. In contrast, prior work such as~\cite{us@BPM22} incorporated selected data attributes into the state, relying on domain knowledge to identify which attributes were relevant. However, in many real-world scenarios, it may be difficult to determine apriori which data attributes are important, and many may be irrelevant to the problem at hand. 
Our approach demonstrates that it is possible to automatically learn effective policies from event logs with minimal domain knowledge---limited to the definition of the optimization goal , i.e., the KPI---even when focusing primarily on control-flow information. This opens the door to future extensions of our methods that can integrate richer process representations, including data or multi-object perspectives, in a similarly automated fashion.

\paragraph{Scope of the Prescription Problem}
Finally, an important conceptual consideration concerns the scope of the prescription problem addressed in this study. Our framework restricts domain knowledge to distinguishing between activities that are controllable by the process owner and those driven by external actors. Within this feasible action space, prescriptions are learned in a fully data-driven manner based on their observed impact on the KPI. While this abstraction enables a general and automated formulation of the prescriptive task, it also implies that the quality and nature of the learned policies depend strongly on the information available in the event log and on the KPI definition. Actions whose effects cannot be inferred from the available state information, or whose contribution is not reflected in the chosen KPI, may not be properly valued by the learned policy.

On the other hand, due to limited state information or scarcity of training data, the learned policy may recommend actions that are not applicable or meaningful in the current context. For instance, an activity may be suggested because it is associated with higher KPI values in the data, even if the conditions required to perform it are not satisfied in the current state. While this issue could be mitigated with richer state representations and sufficient data, it remains a limitation of data-driven policy learning in realistic settings. As a result, learned policies may not fully capture all relevant decision constraints: an agent may recommend actions that are not applicable in the current context, or avoid activities that appear suboptimal under the chosen KPI, even though they may be necessary under alternative objectives.

At the same time, the framework naturally supports settings in which only a subset of activities is controllable, while others are driven by external or exogenous factors. This makes the approach applicable to realistic multi-actor environments, even when full control over the process is not possible.

Consequently, in settings where richer contextual information (e.g., detailed data attributes or risk indicators) or multi-objective KPIs are available, more expressive state representations and reward formulations could support the learning of more nuanced intervention strategies. Exploring these extensions represents an important direction for future work.


\section{Conclusion and Future Directions}
\label{sec:conclusion}

In this paper, we presented and compared two Reinforcement Learning (RL) approaches for learning effective activity recommendation policies from annotated sequences of event data in a multi-actor setting, with the goal of optimizing a given Key Performance Indicator (KPI).
The first approach constructs an explicit Markov Decision Process (MDP) by applying clustering techniques to the event log, followed by the use of a Dynamic Programming algorithm to derive the optimal policy. The second approach leverages Conservative Q-Learning (CQL), a state of the art \odrl method capable of learning policies directly from historical data without requiring interaction with the environment.

Policies learned from data without direct interaction with the system can serve as a strong initialization for more comprehensive optimization frameworks, where the learned policies may be further refined through online interaction or feedback in a subsequent phase.

To evaluate the effectiveness of both approaches, we used Business Process Simulation, testing the policies across a range of synthetic and real-world scenarios.
The results show that both the MDP-based and the \odrl methods are capable of learning robust and effective policies, even in the presence of different log sizes and process characteristics, with the MDP-based approach generally achieving moderately better performance while also being more computationally efficient.

In the future, we plan to address the multi-perspective nature of process mining data. As a first step, we aim to incorporate information from data attributes into the definition of the state space. Subsequently, we intend to extend the action space to include relevant attributes as well. 
For example, in the loan application scenario discussed in this paper, a customer's response to a loan offer may strongly depend on the specific characteristics of the offer itself. For example, some of this information is available in the BPIC2017 dataset and could be used to learn more fine-grained policies. An alternative and complementary direction is the development and use of more expressive synthetic simulation models. In the loan application domain, for instance, such models could incorporate additional aspects such as loan repayment behavior, fraud attempts, or longer-term customer risk profiles. This would enable the evaluation of prescriptive approaches in richer and more realistic settings, where policies must account for longer-term and multi-faceted effects of interventions.

Another interesting direction for future research is the training of context-aware policies that take into account information from concurrent process instances. For example, recommending the same activity to many concurrent cases may saturate the available resources, leading to inefficiencies. 
While in this work we relied on Business Process Simulation, which is a context-aware evaluation technique that ensured our policies did not create such issues, learning context-aware policies directly could further improve overall effectiveness.

\appendix

\section{Sensitivity Analysis on the Number of Clusters}
\label{appx:cluster_number}

\begin{figure}[ht!]
  \centering
  \includegraphics[width=.9\linewidth]{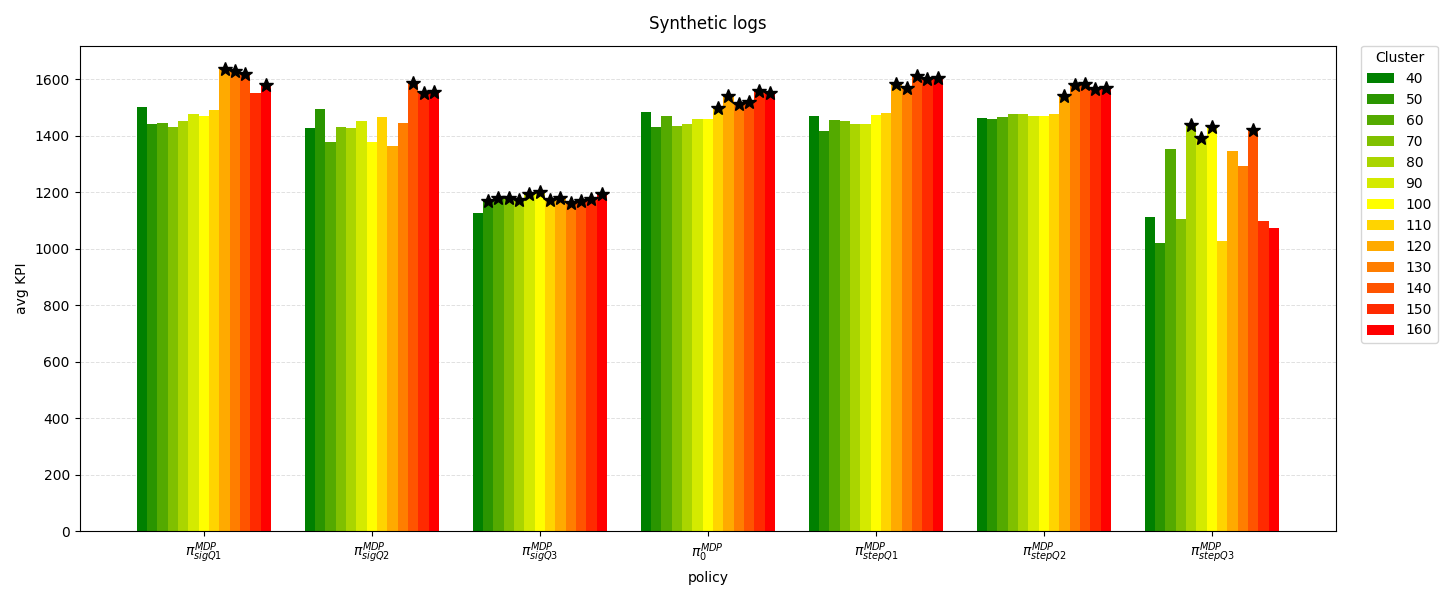}
  \caption{
  Average KPI achieved for each scaling function as the number of clusters varies for the synthetic log $\mathcal{L}_{4000}$. The yellow bar in the center indicates the cluster number selected in the paper.}
  \label{fig:sim_log_cluster}
\end{figure}

\begin{figure}[ht!]
  \centering
  \includegraphics[width=.9\linewidth]{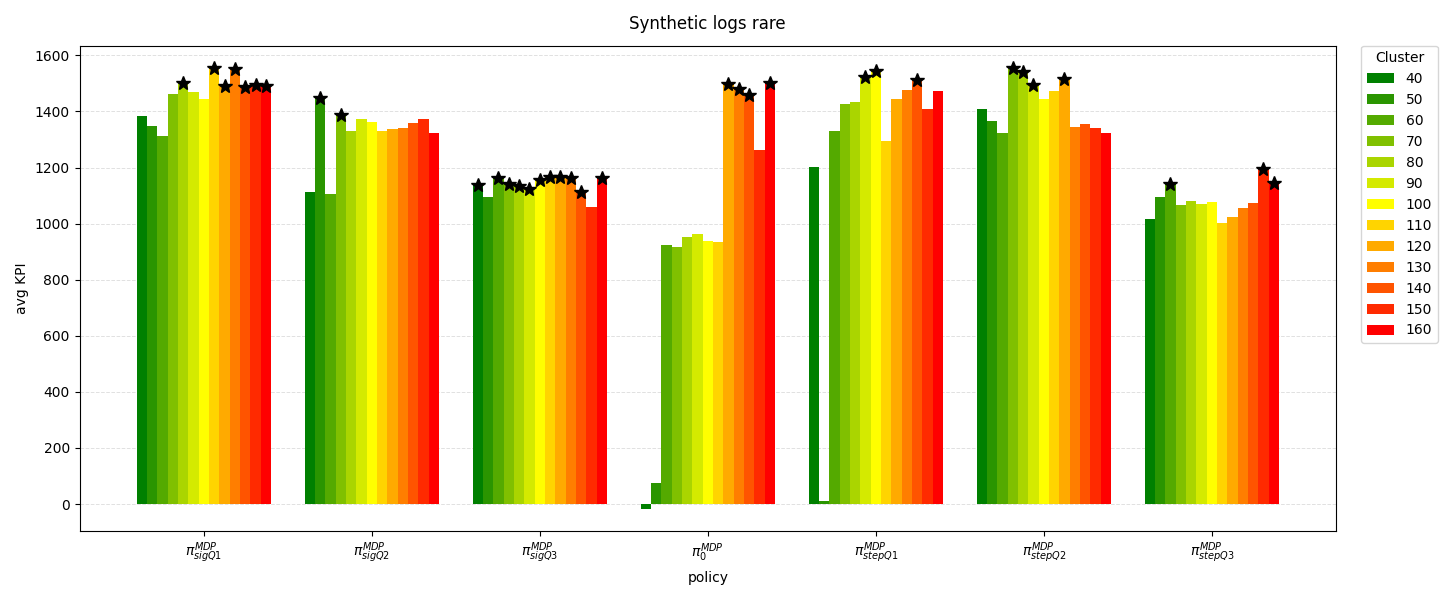}
  \caption{
  Average KPI achieved for each scaling function as the number of clusters varies for the synthetic log $\mathcal{L}_{4000,rare}$. The yellow bar in the center indicates the cluster number selected in the paper.}
  \label{fig:sim_log_rare_cluster}
\end{figure}

We present here an investigation of how the performance of the MDP-based approach depends on the number of clusters used for state abstraction.
Figures~\ref{fig:sim_log_cluster} and~\ref{fig:sim_log_rare_cluster} report the performance of the MDP-based models trained on the synthetic logs $\mathcal{L}_{4000}$ and $\mathcal{L}_{4000,rare}$, respectively, for different choices of the number of clusters used in the state abstraction described in Section~\ref{sec:constructionMDP}.
We consider $13$ different values for the number of clusters, distributed around the reference value adopted in the paper, which was selected using the elbow method on the within-cluster sum of squares (WSS), which is $100$ for both event logs,
and is depicted as yellow bars in the figures.
The results are shown for each scaling function considered in Section~\ref{sec:result}.

These results show that the performance of the learned policies is affected by the number of clusters only for certain scaling function design choices. In particular, only those policies that already exhibited inferior performance in Section~\ref{sec:result} are significantly affected when the number of clusters changes moderately around the reference value of $100$.

This effect is particularly evident for the scaling function $h^{0}$ in the log $\mathcal{L}_{4000,rare}$, which allows infrequently observed behaviors to strongly influence learning, and for $h^\text{sig}_{Q_3}$ in the log $\mathcal{L}_{4000}$, which instead emphasizes only very frequent behaviors.
The introduction of scaling functions was intended precisely to improve training robustness; these results indicate that the aforementioned options do not achieve this goal effectively.

For the more robust scaling functions, namely $h^\text{sig}_{Q_1}$, $h^\text{sig}_{Q_2}$, $h^\text{step}_{Q_1}$, and $h^\text{step}_{Q_2}$, the choice of the number of clusters
has a limited impact on policy performance, at least for moderate variations around the reference value, and can therefore be considered a non-critical design choice,
as clustering mainly serves to provide a reasonable state abstraction that is subsequently exploited during RL training. The stable performance across a wide range of cluster numbers further confirms the robustness of these scaling functions and supports their suitability as default choices in our setting.

\section{Synthetic Evaluation of Later Recommendations}
\label{appx:simulation}

We present here a more detailed evaluation of the policies in the synthetic scenarios considered in Section~\ref{sec:answer_RQ1}. We focus on the effectiveness of each policy when the agent starts recommending the next activity at a later stage in the process execution, after a number of activities have already been carried out using the customary policy. The customary policy refers to the default policy used by the simulation model $\mathcal{M}$ to generate the synthetic datasets, representing a predefined, non-optimized approach.
To perform this evaluation, we start the simulation of a trace using its built-in policy for a fixed number of events. After that, we start selecting the next activities following the agent's recommendations and compute the future KPI gathered from that point onward, ignoring the contribution of previous activities. In this analysis, we consider two learned policies: $\pi^\text{MDP}_{\text{sig}Q_1}$, from the MDP-based approach, and $\pi^{\text{CQL}}_{100}$, from the Offline Deep RL approach, and we compare them with the informed heuristic baseline $\pi_\text{informed}$.
We consider seven different starting moments for the agent: after $5$, $10$, $15$, $20$, and $25$ events have occurred, respectively. At $0$, we include the evaluation of the agent from Section~\ref{sec:answer_RQ1}, where it is active from the beginning. Each data point corresponds to a simulation of $2000$ traces.

\begin{figure}[ht!]
  \centering
  \includegraphics[width=1\linewidth]{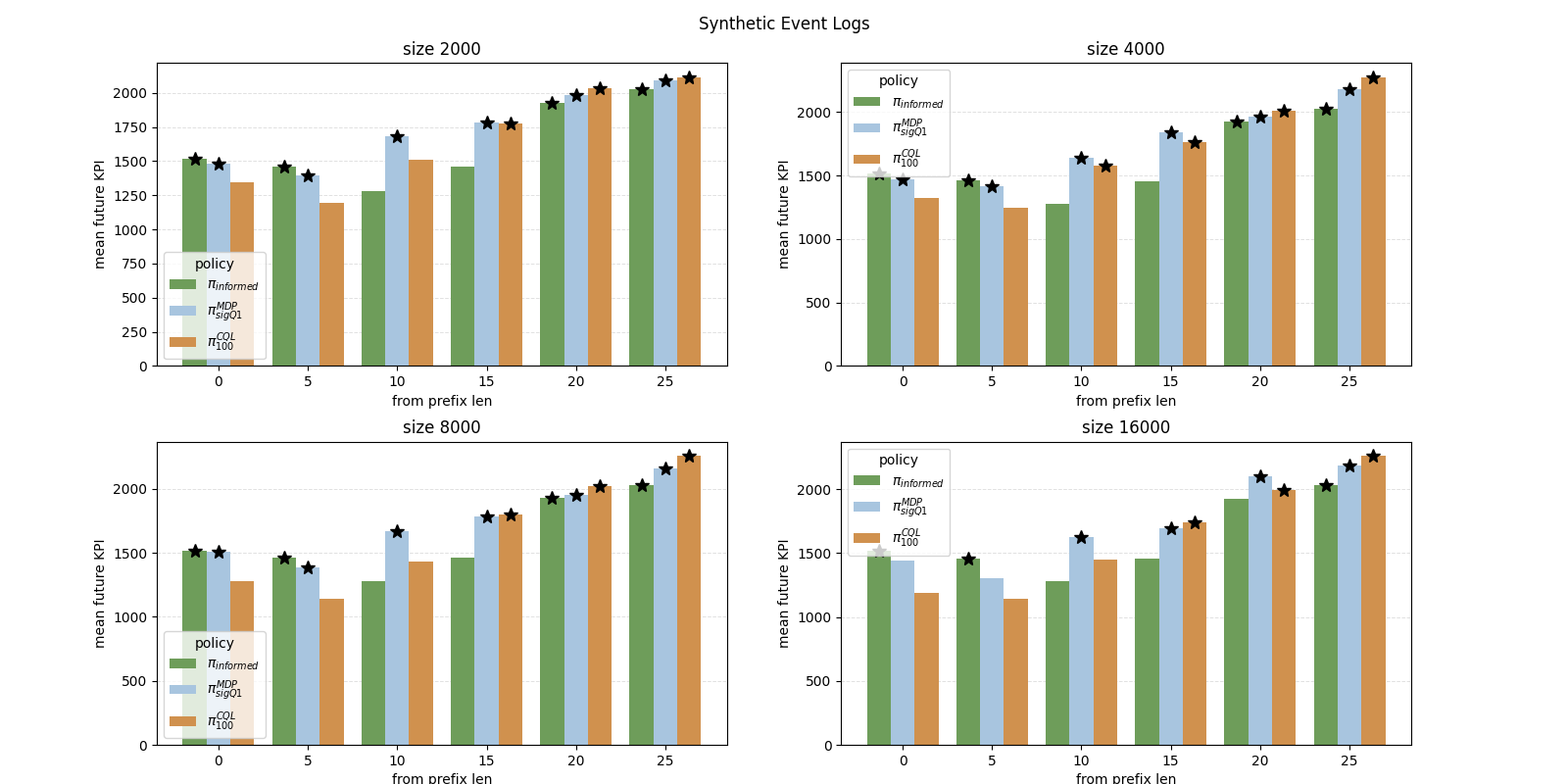}
  \caption{Average KPI achieved when policies are applied starting from a given point in the trace execution, for synthetic logs $\mathcal{L}_{size}$.}
  \label{fig:sim_log_evolution}
\end{figure}

\begin{figure}[ht!]
  \centering
  \includegraphics[width=1\linewidth]{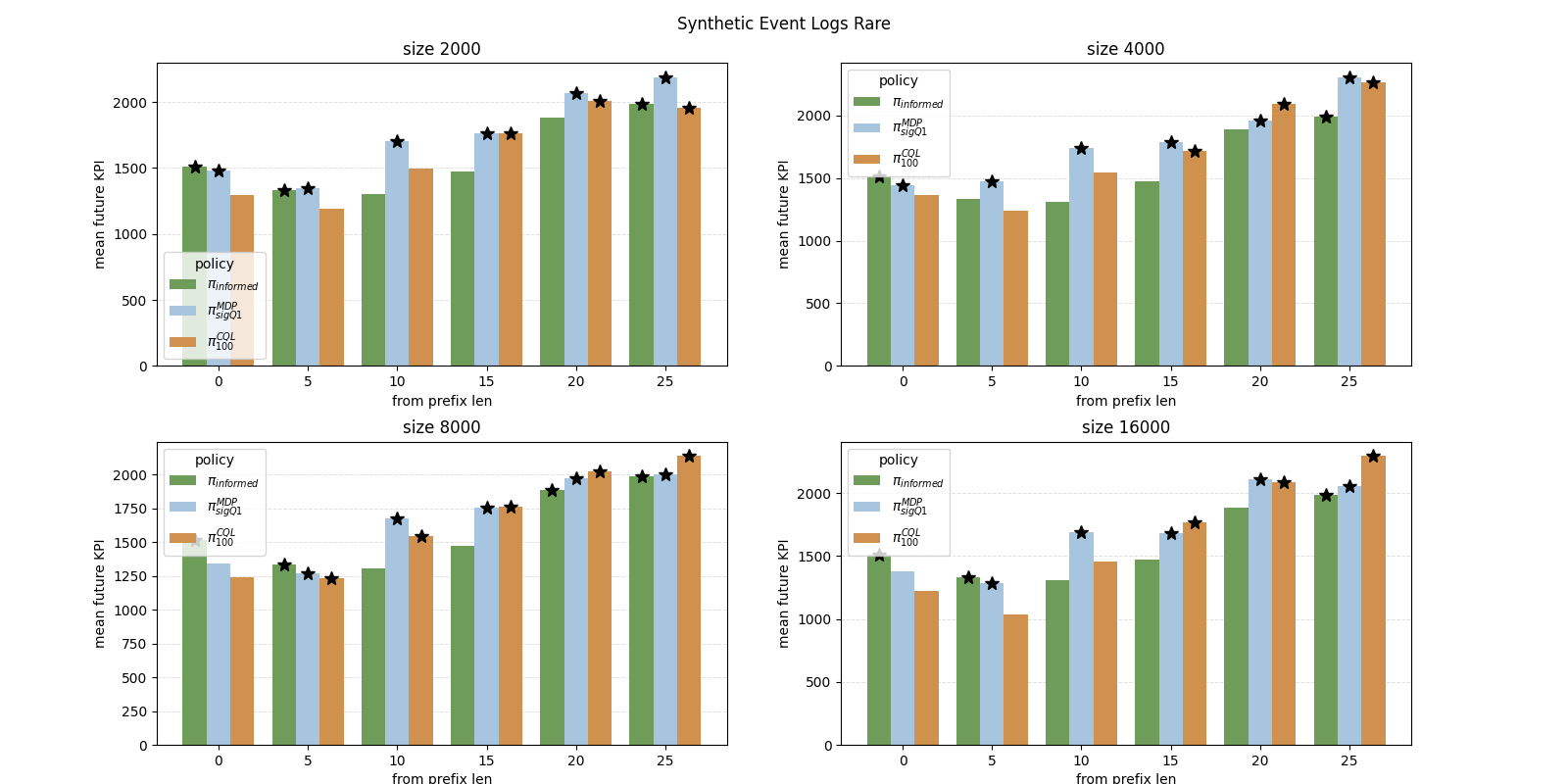}
  \caption{Average KPI achieved when policies are applied starting from a given point in the trace execution, for synthetic logs $\mathcal{L}_{size,rare}$.}
  \label{fig:sim_log_rare_evolution}
\end{figure}

In Figure~\ref{fig:sim_log_evolution} and Figure~\ref{fig:sim_log_rare_evolution}, the results corresponding to the $\mathcal{L}_{size}$ and $\mathcal{L}_{size,rare}$ synthetic logs are shown, respectively.
For each log and prefix length, the best performing policy, as well as those whose performance is not statistically significantly different from the best, are marked with a black star above the corresponding bar.

First, we observe that the informed heuristic policy, which in the evaluation of Section~\ref{sec:answer_RQ1} consistently achieved top performance, is now outperformed by the learned policies in some cases when the prefix length is between 5 and 20, and consistently loses to at least one of the two learned policies across all logs at prefix lengths 10 and 15. This suggests that, being a fixed heuristic, it may fail to adapt to specific situations compared to the learned policies.
In contrast, the MDP-based policy $\pi^\text{MDP}_{\text{sig}Q_1}$ emerges as the most robust, consistently ranking among the best-performing policies, with only four exceptions, in which it is still not outperformed by the \odrl policy.
Finally, we observe that the \odrl policy, which is consistently outperformed when applied from the start of the execution, becomes more effective at later stages, consistently appearing among the best-performing policies for prefix lengths 15, 20, and 25.


\section{Business Process Simulation Model of Synthetic Logs}
\label{appx:simulation_2}

\begin{figure}[ht!]
  \centering
  \includegraphics[angle=90, width=0.28\textwidth]{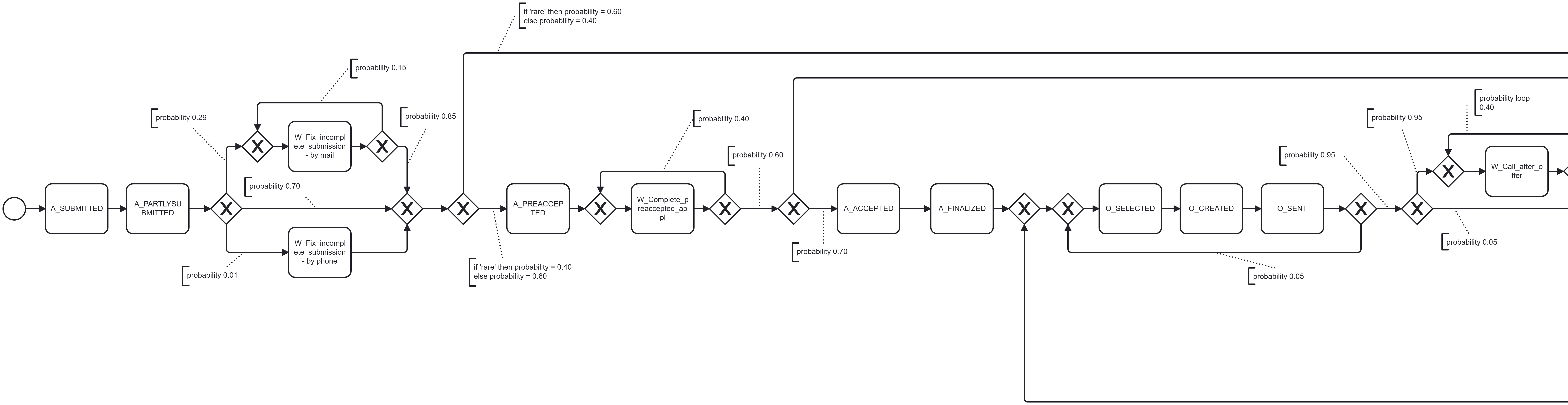}
  \hspace{0.1\textwidth} 
  \includegraphics[angle=90, width=0.28\textwidth]{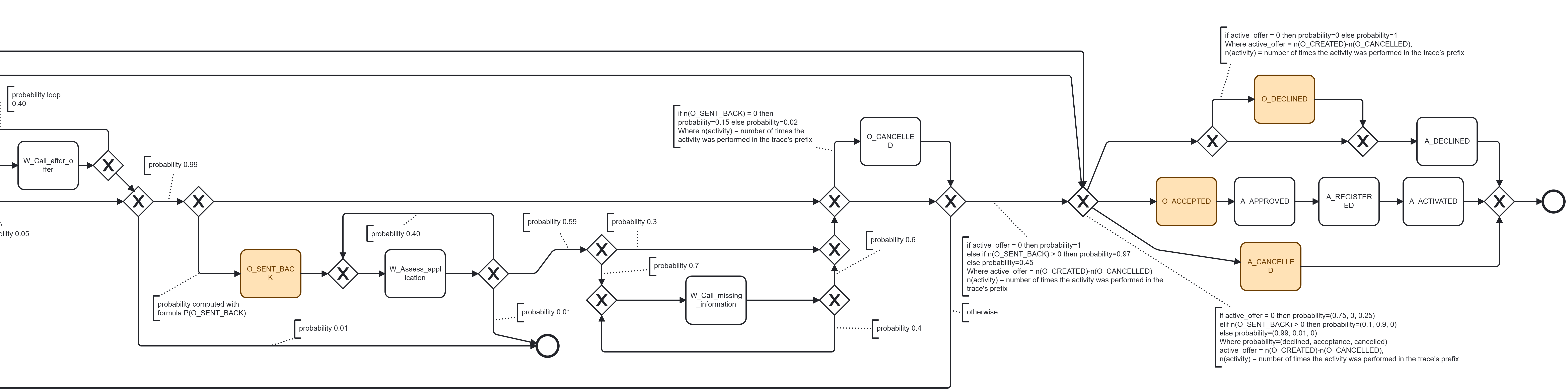}
  \caption{The BPMN model incorporated into the simulation model $\mathcal{M}$ for generating the eight synthetic logs, with conditional probabilities specified at each decision point based on the success rate ($rare$). Activities executed by the customer are highlighted in orange. For clarity of visualization, the model is shown split into two parts.}
  \label{fig:complete_BPMN_syn}
\end{figure}

Figure~\ref{fig:complete_BPMN_syn} shows the complete BPMN model incorporated as $\mathcal{N}$ in the simulation model $\mathcal{M}$, which is used to generate the synthetic logs described in Table~\ref{tab:dataset}. Figure~\ref{fig:complete_BPMN_syn} also presents the $\mathcal{P}_C$ simulation parameters defined at each decision point in $\mathcal{N}$, which govern the selection between alternative paths during simulation.

Activities executed by the customer are highlighted in orange. One of the most critical customer's behavior influencing the outcome of the process is the probability to perform the activity \actlabel{O\_SENT\_BACK} which in our model is governed as described in the following. 
We start from a baseline probability $P(\actlabel{O\_SENT\_BACK}) = 0.6$, which is adjusted according to the rules reported in Table~\ref{tab:o_sent_back_prob}.

\begin{table}[ht!]
\centering
\resizebox{0.8\textwidth}{!}{%
\begin{tabular}{llc}
\toprule
\textbf{Attribute} & \textbf{Condition} & $\boldsymbol{\Delta P (\%)}$ \\
\midrule
Amount
& $\le 10000$ & $+10$ \\
& $> 10000$ & $-20$ \\
\midrule
$n(\actlabel{O\_CREATED})$
& $= 0$ & $-100$ \\
& $= 1$ & $0$ \\
& $= 2$ & $+10$ \\
& $= 3$ & $+20$ \\
& $= 4$ & $0$ \\
& $= 5$ & $-20$ \\
& $> 5$ & $-40$ \\
\midrule
Active offers 
& $= 0$ & $-100$ \\
$= n(\actlabel{O\_CREATED}) - n(\actlabel{O\_CANCELLED})$
& $= 1$ & $0$ \\
& $= 2$ & $+40$ \\
& $= 3$ & $0$ \\
& $> 3$ & $-40$ \\
\midrule
$n(\actlabel{W\_Call\_after\_offer})$
& $= 0$ & $-10$ \\
& $\in [1,3] \cup [7,10]$ & $0$ \\
& $\in [4,6]$ & $+10$ \\
& $> 10$ & $-30$ \\
\midrule
$n(\actlabel{W\_Call\_missing\_information})$
& $= 0$ & $0$ \\
& $\in [1,5]$ & $-5$ \\
& $\in [6,10]$ & $-10$ \\
& $> 10$ & $-30$ \\
\midrule
$n(\actlabel{O\_SENT\_BACK})$
& $= 0$ & $0$ \\
& $> 0$ & $-60$ \\
\bottomrule
\end{tabular}
}
\caption{Adjustments to $P(\actlabel{O\_SENT\_BACK})$ based on the number of occurrences of key activities in the trace.}
\label{tab:o_sent_back_prob}
\end{table}

\subsection{Informed Heuristic Policy Definition}
\label{appx:baseline_policy}

We leverage the knowledge of the simulation logic described in the previous section to define a meaningful policy. This policy is a simple heuristic aimed at maximizing the probability of the activity \actlabel{O\_SENT\_BACK}, as detailed in Table~\ref{tab:o_sent_back_prob}. In fact, the execution of this activity is highly correlated with the occurrence of \actlabel{O\_ACCEPTED}, which defines the positive outcome of a process execution.

The policy is implemented as a hard-coded, rule-based strategy detailed in Table~\ref{tab:baseline_rules}, which selects the next activity based on the last executed activity and the frequency of key activities in the current trace execution.

\begin{table}[ht!]
\centering
\resizebox{\textwidth}{!}{%
\begin{tabular}{lll}
\toprule
\textbf{Current Activity} & \textbf{Condition} & \textbf{Next Activity} \\
\midrule
\actlabel{A\_SUBMITTED} & - & \actlabel{A\_PARTLYSUBMITTED} \\
\actlabel{A\_PARTLYSUBMITTED} & - & \actlabel{A\_PREACCEPTED} \\
\actlabel{W\_Fix\_incomplete\_submission\_by\_mail} & - & \actlabel{A\_PREACCEPTED} \\
\actlabel{W\_Fix\_incomplete\_submission\_by\_phone} & - & \actlabel{A\_PREACCEPTED} \\
\actlabel{A\_PREACCEPTED} & - & \actlabel{W\_Complete\_preaccepted\_appl} \\
\actlabel{W\_Complete\_preaccepted\_appl} & - & \actlabel{A\_ACCEPTED} \\
\actlabel{A\_ACCEPTED} & - & \actlabel{A\_FINALIZED} \\
\actlabel{A\_FINALIZED} & - & \actlabel{O\_SELECTED} \\
\actlabel{O\_SELECTED} & - & \actlabel{O\_CREATED} \\
\actlabel{O\_CREATED} & - & \actlabel{O\_SENT} \\
\actlabel{O\_SENT\_BACK} & - & \actlabel{W\_Assess\_application} \\
\actlabel{O\_DECLINED} & - & \actlabel{A\_DECLINED} \\
\actlabel{O\_ACCEPTED} & - & \actlabel{A\_APPROVED} \\
\actlabel{A\_APPROVED} & - & \actlabel{A\_REGISTERED} \\
\actlabel{A\_REGISTERED} & - & \actlabel{A\_ACTIVATED} \\
\midrule
\multirow{2}{*}{\actlabel{O\_SENT}} 
  & if $\text{Active offers} < 2$ & \actlabel{O\_SELECTED} \\
  & else & \actlabel{\actlabel{W\_Call\_after\_offer}} \\
\midrule
\multirow{3}{*}{\actlabel{W\_Call\_after\_offer}} 
  & if $n(\actlabel{W\_Call\_after\_offer}) \le 6$ & \actlabel{W\_Call\_after\_offer} \\
  & else if $n(\actlabel{O\_CREATED}) < 3$ & \actlabel{O\_CANCELLED} \\
  & else & \actlabel{A\_DECLINED} \\
\midrule
\multirow{2}{*}{\actlabel{O\_CANCELLED}} 
  & if $n(\actlabel{O\_CREATED}) < 3$ & \actlabel{O\_SELECTED} \\
  & else & \actlabel{A\_DECLINED} \\
\midrule
\multirow{2}{*}{\actlabel{W\_Assess\_application}}
  & if $n(\actlabel{O\_CREATED}) < 3$ & \actlabel{O\_CANCELLED} \\
  & else & \actlabel{A\_DECLINED} \\
\midrule
\multirow{2}{*}{\actlabel{W\_Call\_missing\_information}} 
  & if $n(\actlabel{O\_CREATED}) < 3$ & \actlabel{O\_CANCELLED} \\
  & else & \actlabel{A\_DECLINED} \\
\bottomrule
\end{tabular}
}
\caption{Informed heuristic policy $\pi_\text{informed}$ for activity selection. Conditional rules depend on the number of occurrences of key activities in the trace.}
\label{tab:baseline_rules}
\end{table}

\clearpage


\section*{Declaration of generative AI and AI-assisted technologies in the manuscript preparation process}
During the preparation of this work, the authors used Open AI\footnote{\url{https://openai.com/}} tools to assist with language editing and improving the clarity of the manuscript. After using these tools, the authors reviewed and edited the content as needed and take full responsibility for the content of the published article.

\bibliographystyle{elsarticle-num} 
\bibliography{bibliography}

\end{document}